\documentclass[11pt]{article}
\usepackage[utf8]{inputenc}
\usepackage[margin=1in]{geometry}
\usepackage[T1]{fontenc}
\usepackage{lmodern}
\usepackage{microtype}
\usepackage{graphicx}
\usepackage{subfigure}
\usepackage{booktabs} 
\usepackage{natbib}
\setcitestyle{open={(},close={)}}
\usepackage{algorithm2e}
\usepackage{amsmath}
\usepackage{mathtools}
\usepackage{amsthm}
\usepackage{graphicx}
\usepackage{textcomp}
\usepackage{xcolor}
\usepackage{amsfonts}     
\usepackage{epsfig}
\usepackage{wrapfig}
\usepackage{balance}
\usepackage{url}
\usepackage{mathtools}
\usepackage{natbib}
\usepackage{multirow}
\SetKwInput{KwInput}{Input}
\SetKwInput{KwData}{Data}
\SetKwInput{KwResult}{Result}

\DeclareMathOperator*{\argmin}{arg\,min}

\usepackage[textsize=tiny]{todonotes}

\usepackage[colorlinks=true,citecolor=blue]{hyperref}

\usepackage{amsmath,amsthm,amsfonts,amssymb,mathdots,array,mathrsfs,bm,bbm,stmaryrd,graphicx,subfigure,xcolor}

\usepackage{breakcites}

\usepackage[T1]{fontenc}
\usepackage{enumerate}
\usepackage{inputenc}

\usepackage{graphicx} 
\usepackage{subfigure}

\usepackage{booktabs,balance}
\usepackage{rotating}
\usepackage{boldline}
\usepackage{makecell}
\usepackage{multirow}
\usepackage{balance}

\usepackage{tikz}

\newtheorem{theorem}{Theorem}[section]

\newtheorem{lemma}[theorem]{Lemma}

\newtheorem{definition}{Definition}[section]

\newtheorem{remark}{Remark}[section]

\newcommand{\bsmat}{\begin{bmatrix} }
\newcommand{\esmat}{\end{bmatrix} }

\usepackage{natbib}
\setcitestyle{numbers}
\setcitestyle{square}

\begin{document}

\title{\bf\Huge Dynamic Incremental Optimization for Best Subset Selection}

\author{\vspace{0.1in}\\\textbf{Shaogang Ren} \vspace{0.05in}  \\ 
shaogang-ren@utc.edu\\
Department of Computer Science and  Engineering\\
University of Tennessee at Chattanooga\\\\
\textbf{Xiaoning Qian} \vspace{0.05in}  \\
Department of Electrical and Computer Engineering\\
Texas A\&M University\\
}

\date{\vspace{0.5in}}
\maketitle

\begin{abstract}\vspace{0.3in}
\vspace{0.1in}
Best subset selection is considered the `gold standard' for many sparse learning problems. A variety of optimization techniques have been proposed to attack this non-smooth non-convex problem.  In this paper, we investigate the dual forms of a family of $\ell_0$-regularized problems. An efficient primal-dual algorithm is developed based on the primal and dual problem structures. By leveraging the dual range estimation along with the incremental strategy, our algorithm  potentially reduces redundant computation and improves the solutions of best subset selection. Theoretical analysis and experiments on synthetic and real-world datasets validate the efficiency and statistical properties of the proposed solutions.
\end{abstract}

\section{Introduction}

Sparse learning is a standard approach to alleviate model over-fitting issues when the feature dimension is larger than the number of training samples. With a training set $\{x_i, y_i\}_{i=1}^n$ where $x_i \in \mathbb{R}^p $ is the sample feature and $y_i$ is the corresponding label, this paper focuses on the following generalized best subset selection problem,
\begin{align}\label{eq:primal1}
&\min_{\beta \in \mathbb{R}^p} P(\beta) = f(\beta) + \lambda_0 ||\beta||_0,  \\\notag
\text{where }\hspace{0.2in} &f(\beta) =   \sum_{i=1}^n l(\beta^{\top} x_i, y_i)   +  \lambda_1 ||\beta||_1 +\lambda_2 ||\beta||_2^2.
\end{align}
Here $l(\cdot)$ is a convex function, $\beta$ is the model parameter,  and $\lambda_0$,  $\lambda_1$ and  $\lambda_2$ are hyper-parameters/tuning parameters. It is well-known that an $\ell_0$ solver ($\lambda_0>0,\lambda_1=0$) has superior statistical properties when the \textit{signal-to-noise ratio} (SNR) is high, but it may suffer from over-fitting issues when SNR is low~\citep{pmlr-v65-david17a,mazumder2022subset}. The continuous-shrinkage solvers  e.g., ridge/LASSO ($\lambda_0=0,\lambda_1>0$), can perform better in this case compared with $\ell_0$ solver~\citep{mazumder2022subset,hastie2017extended}. Combinations of these hyper-parameters may adjust the model to work well in different noise levels. \cite{mazumder2022subset,Hazimeh18} used ridge/LASSO to improve $\ell_0$ solutions and achieve better or comparable solutions~with~less~nonzeros.

 By leveraging the significant computational advances in mixed-integer optimization (MIO), \cite{Bertsimas2015BestSS} performed near optimal solutions to a special case of problem~\eqref{eq:primal1}, for $\lambda_1, \lambda_2 = 0$. This method scales up solutions to  cases where feature sizes are much larger than what were considered possible in the community~\citep{Furnival74,Hazimeh18}. Their approach can achieve approximate optimality via dual bounds but with the cost of longer computation time. \cite{Bertsimas17} showed that cutting plane methods for subset selection can work well with mild sample correlations and a succinctly large $n$.

 Different from the soft regularized ridge/LASSO problem given by~\eqref{eq:primal1}, Iterative Hard Thresholding~(IHT)~\citep{blumensath2009iterative,foucart2011hard,yuan2014gradient,jie2017tight,yuan2020nearly,Liu17,yuan2020dual} has often been used to solve  $k$-sparse problems~\eqref{eq:hard_k}, 
\begin{align}\label{eq:hard_k}
&\min_{||\beta||_0 \leq k} \sum_{i=1}^n l(\beta^{\top} x_i, y_i)    +\lambda_2 ||\beta||_2^2 \ .
\end{align}
In~\cite{blumensath2009iterative,foucart2011hard}, the authors demonstrated that IHT can be applied to compute the compressed sensing problem. IHT-based approaches have been studied by many researchers in the context of sparse learning problems~\citep{yuan2014gradient,jain2014iterative,jain2016structured,yuan2020nearly}. IHT methods require a specific value of the feature number~($k$) to start the algorithm. Apart from IHT, many different approaches have also been developed to tackle the $\ell_0$ regularized problems~\citep{Bertsimas2015BestSS,Mazumder15,mazumder2022subset,soussen2015homotopy,bian2020smoothing,dedieu2020learning,yang2020fast,dong2015regularization,hazimeh2020sparse,zhu2020polynomial}.

Apart from $\ell_0$ solvers, extremely efficient and optimized   $\ell_1$-regularization (LASSO) solvers can solve an entire regularization path (with a hundred values of the tuning parameter) in usually less than a second~\citep{friedman2010regularization}. Screening and coordinate incremental techniques~\citep{Fercoq2015,GAP,ren2017scalable,ren2020thunder} can further scale the solutions to large datasets.  Compared to popular efficient solvers for LASSO, it seems that the high computation cost for using $\ell_0$ regularized models ~\citep{Bertsimas2015BestSS} might discourage practitioners from adopting global optimization-based solvers of~\eqref{eq:primal1} to daily analysis applications~\citep{hastie2017extended,mazumder2022subset,Hazimeh18}.  However, it is known~\citep{Loh14,Hazimeh18} that there is a significant gap in the statistical quality of solutions that can be achieved via LASSO (and its variants) and near-optimal solutions to non-convex subset-selection type procedures. The choice of algorithm can significantly affect the quality of solutions obtained. On many instances, algorithms that~do a better job in optimizing the non-convex subset-selection criterion ~\eqref{eq:primal1} result in superior-quality statistical estimators (for example, in terms of support recovery~\citep{Hazimeh18}).

Several recent studies attempt to further improve the efficiency of $\ell_0$ solvers. Along the line of dual methods, ~\cite{Liu17} and~\cite{yuan2020dual}  studied the strong duality  of $k$-sparse problem, and they proposed  dual space hard-thresholding methods that attain optimal solutions within polynomial computation complexity. Besides dual-based methods, screening methodology  has been extended to $\ell_0$ solvers by researchers~\cite{atamturk2020safe}. Following coordinate descent~(CD) methods~\citep{breheny2011coordinate,mazumder2011sparsenet,friedman2010regularization,nesterov2012efficiency} for linear regression problems,~\cite{Hazimeh18} proposed an efficient CD-based method to scale up the solutions of problem~\eqref{eq:primal1}. Their method can be improved with the proposed switch techniques that aim to escape from local solutions. 
Additionally, the combination of $\ell_0$ and $\ell_q$ ($q = 1$ or 2) is considered in  the existing literature. See~\cite{liu2007variable, soubies2017unified} for theoretical analyses and details.

Following the studies in~\cite{pilanci2015sparse,Liu17,yuan2020dual},  we investigate the dual form of the generalized sparse problem~\eqref{eq:primal1}. Under mild conditions, a strong duality theory can be established for problem~\eqref{eq:primal1}. A primal-dual algorithm~is proposed to further improve the efficiency and quality of solutions by leveraging the exploration in the~dual space along with coordinate screening and active incremental techniques~\citep{Fercoq2015,GAP,Ndiaye2017,atamturk2020safe,Celer,ren2017scalable,ren2020thunder}. \textbf{Our contributions} on the theoretical side of best subset selection are three-fold: (1) We derive the dual form of the generalized non-convex sparse learning problem~\eqref{eq:primal1}. (2) We demonstrate that the derived strong duality allows us to adopt the screening and coordinate incremental strategies~\citep{Fercoq2015,GAP,Ndiaye2017,ren2017scalable,ren2020thunder} in $\ell_1$ solvers to boost the efficiency of the proposed  $\ell_0$ algorithm. (3) We provide theoretical analyses of the proposed algorithms, which show that the generalized sparse problem~\eqref{eq:primal1} can be solved with polynomial  complexity. 
Experiments on both synthetic and real-world datasets show the advantages of our method.

The rest of the paper is organized as follows. In Section~\ref{sec:duality}, we formulate the dual form of the generalized sparse learning problem.
In Section~\ref{sec:algorithm}, we propose the new primal-dual algorithm improved with coordinate incremental techniques. Section~\ref{sec:analysis} presents our algorithm analyses. 
Experimental results are provided in Section~\ref{sec:experiment}.
A discussion is given in Section~\ref{sec:discussion}, and the concluding remark is presented in Section~\ref{sec:conclusion}.

\vspace{0.2in}

\noindent \textbf{Notation.}\ 
Symbol $\beta \in \mathbb R^p$ is used for the primal variable and $\alpha \in \mathbb R^n$ is for the dual variable. 
We use $\|\beta\|$, $\|\beta\|_0$ and $\|\beta\|_1$ to denote the $\ell_2$, $\ell_0$ and $\ell_1$ norm of $\beta$, respectively.
Functions $P(\beta)$ and $D(\alpha)$ represent the primal objective and the dual objective correspondingly. For matrix $X$, $\sigma_{max}(X)$/$\sigma_{min}(X)$ denotes its largest/smallest singular value. $\mathrm{supp}(\beta)$ is the support set of vector $\beta$, i.e. $\mathrm{supp}(\beta)=\{j | \beta_j \neq 0\}$. $S^c$ represents the complement of set $S$.  

\section{Properties of Generalized Sparse Learning}\label{sec:duality} 
This section extends the duality studies  in~\cite{pilanci2015sparse,Liu17,yuan2020dual} to the generalized sparse learning problem~\eqref{eq:primal1}. The dual problem of~\eqref{eq:primal1} is introduced in Section~\ref{sec:dual}, and then the range estimation of the dual variable using the duality gap is derived in Section~\ref{sec:dual_v_est}.

\vspace{-0.05in}
\subsection{Dual Problem}\label{sec:dual}
\vspace{-0.05in}

Let $X =[x_1, x_2, ..., x_n]^{\top}$ be the feature matrix, $y= [y_1, y_2, ..., y_n]^{\top}$ be the response vector and $n$ is the number of samples. 
Let $l^*_i(\alpha_i) = \max_{u\in \mathcal{F}} \{\alpha_i u - l_i(u)\}$ be the Fenchel conjugate~\citep{fenchel1949conjugate} of the convex loss function $l_i(u)$ and $\mathcal{F} \subseteq \mathbb{R} $ be the feasible set of $\alpha_i$ regarding $l^*_i()$. According to the expression $l_i(u) = \max_{\alpha_i \in \mathcal{F}}\{\alpha_i u - l_i^*(\alpha_i)\}$, the primal problem can be reformulated as 
\begin{align} \label{eq:Lminmax}
\min_{\beta} \sum_{i=1}^n \max_{\alpha_i \in \mathcal{F}} &\big(\alpha_i \beta^{\top}x_i - l_i^*(\alpha_i) \big) + \lambda_0 ||\beta||_0    +  \lambda_1 ||\beta||_1 + \lambda_2 ||\beta||^2.
 \end{align}
We use $L(\beta, \alpha)$ to represent the following objective
\begin{eqnarray} ~\label{eq:lagrangian}
L(\beta, \alpha) =  & \sum_{i=1}^n \big(\alpha_i \beta^{\top}x_i - l_i^*(\alpha_i) \big) + \lambda_0 ||\beta||_0    +  \lambda_1 ||\beta||_1 + \lambda_2 ||\beta||^2 .
 \end{eqnarray}

Similar to the studies in~\cite{Liu17,yuan2020dual}, the  RIP (restricted isometry property strong condition number) bound conditions are not explicitly required here. Without specifying $k$ in~\eqref{eq:hard_k}, our duality theory is close to the standard duality paradigm. We define the saddle point for the Lagrangian~\eqref{eq:lagrangian} of the generalized sparse learning~\eqref{eq:primal1}. 
\begin{definition} 
(Saddle Point). A pair $(\bar{\beta}, \bar{\alpha} ) \in \mathbb{R}^p \times \mathcal{F}^n$ is said to be a  saddle point for $L$~\eqref{eq:lagrangian}  if  the following holds 
 \begin{align}
L(\bar{\beta}, \alpha) \leq L(\bar{\beta}, \bar{\alpha})  \leq L( \beta, \bar{\alpha}), \  \   \forall  (\beta, \alpha) \in \mathbb{R}^p \times \mathcal{F}^n. \label{eq:saddlepoint}
\end{align}
\end{definition}

Given $\alpha \in \mathcal{F}^n$, we  define  $\eta(\alpha) := - \frac{1}{2\lambda_2} \sum_{i=1}^n \alpha_i x_i  = - \frac{1}{2\lambda_2}  X^{\top}\alpha $, $\eta_0 := \frac{ 2\sqrt{\lambda_0 \lambda_2} +  \lambda_1}{2\lambda_2}$, and 
\begin{align} \label{eq:B_frak}
\beta_j(\alpha):=  \mathfrak{B}(\eta_j(\alpha))  :=  \begin{cases}
\mathrm{sign}\big(\eta_j(\alpha)\big) \big(|\eta_j(\alpha)| -\frac{\lambda_1 }{2\lambda_2} \big) &\text{if} \  \    |\eta_j(\alpha)|  > \eta_0 \\
\big\{0, \mathrm{sign}\big(\eta_j(\alpha)\big) \big(|\eta_j(\alpha)| -\frac{\lambda_1 }{2\lambda_2} \big) \big\}  &\text{if}  \  \  |\eta_j(\alpha)|  = \eta_0  \\
0  &\text{if}  \  \   |\eta_j(\alpha)|  < \eta_0
\end{cases}
.
\end{align}
Moreover, we define 
\begin{align}\label{eq:Psi_1}
&\Psi(\eta_j(\alpha); \lambda_0,  \lambda_1, \lambda_2)   
:=  \begin{cases}
- \lambda_2 (|\eta_j(\alpha)| - \frac{\lambda_1}{2 \lambda_2})^2 + \lambda_0   &\text{if} \  \     |\eta_j(\alpha)|  > \eta_0  \\
\big\{0, - \lambda_2 (|\eta_j(\alpha)| - \frac{\lambda_1}{2 \lambda_2})^2 + \lambda_0  \big\} \quad  &\text{if}  \  \    |\eta_j(\alpha)|  = \eta_0 \\
0  &\text{if}  \  \   |\eta_j(\alpha)|  < \eta_0
\end{cases} .
\end{align}
Then, the corresponding dual problem of~\eqref{eq:primal1} is  written as
\begin{align} \label{eq:dual}
  &\max_{\alpha \in \mathcal{F}^n} D(\alpha) =  \max_{\alpha \in \mathcal{F}^n}-  \sum_{i=1}^n  l_i^*(\alpha_i)   + \sum_{j=1}^p \Psi(\eta_j(\alpha); \lambda_0,  \lambda_1, \lambda_2),
\end{align} 
where  $l^*$ is the conjugate function of $l$.  The primal dual link is written as $\beta_j(\alpha)  =  \mathfrak{B}(\eta_j(\alpha))$.  Here $\eta_0$ is the threshold that controls the sparsity of the solution. Equation~\eqref{eq:B_frak} returns the primal solution given a dual problem solution.  A more detailed derivation of the dual form~\eqref{eq:dual} and strong duality is provided in the supplemental file.

\vspace{-0.05in}
\subsection{Dual Variable Estimation}\label{sec:dual_v_est}
\vspace{-0.05in}
In this paper, we study the duality of the generalized sparse learning problem. Based on the strong duality of $\ell_0$ problem~\eqref{eq:primal1}, screening methods~\citep{Fercoq2015,GAP,Ndiaye2017} and coordinate increasing techniques~\citep{ren2017scalable,ren2020thunder} can be implemented to improve  algorithm efficiency. Following the derivation of the GAP screening algorithm~\citep{Fercoq2015,GAP} for the LASSO problem, we have the following theorem regarding the duality gap. 

\begin{theorem}\label{Thm:ball}
Assume that the primal loss functions $\{l_i(\cdot)\}_{i=1}^n$ are $1/\mu$-strongly smooth.  The range of the dual variable is  bounded via the duality gap value, i.e.,  $\forall \alpha \in \mathcal{F}^n, \beta \in \mathbb{R}^p$, $\{B(\alpha; r) : || \alpha -  \bar{\alpha}||_2  \leq r, r = \sqrt{ \frac{2(P(\beta) - D(\alpha)) } {\gamma}} \} $. 
Here $\gamma$ is a positive constant and $\gamma \geq \mu$. 
\end{theorem}

Let $x_{\cdot i}$ be the $i$th column of $X$, according to the definition of $\mathfrak{B}$ in~\eqref{eq:B_frak}, the activity of feature $j$ is determined by the magnitude of $\eta_j(\bar{\alpha})$, i.e.,  $|\eta_j(\bar{\alpha})| = \frac{1}{2\lambda_2} |x^{\top}_{\cdot j} \bar{\alpha}|$.  With the ball region estimation for $\bar{\alpha}$ in Theorem~\ref{Thm:ball}, we can estimate the activity of a feature with the value of current $\alpha$. Let  $r= \sqrt{ \frac{2(P(\beta) - D(\alpha)) } {\gamma}}$ be the radius of the estimated ball range for $\bar{\alpha}$ using current $\alpha$ and $\beta$ solutions. Then 
$\big||x^{\top}_{\cdot j} \alpha| - ||x_{\cdot j}||_2 r\big| \leq |x^{\top}_{\cdot j} \bar{\alpha}| \leq |x^{\top}_{\cdot j} \alpha| + ||x_{\cdot j} ||_2 r $ and
 we get $|\eta_i(\bar{\alpha})| \leq  \frac{1}{2\lambda_2}\big(|x^{\top}_{\cdot j} \alpha| + ||x_{\cdot j} ||_2 r\big) <\eta_0  \implies x_{\cdot j}  \  \text{is an inactive feature}.$
It implies
\begin{align*}
&\text{Upper Bound}: |x^{\top}_{\cdot j} \alpha| + ||x_{\cdot j} ||_2 r  <2\lambda_2 \eta_0 = 2\sqrt{\lambda_0 \lambda_2} +  \lambda_1   \implies j \notin \mathrm{supp}(\bar{\beta}), \\
&\text{Lower Bound}: \big| |x^{\top}_{\cdot j} \alpha| - ||x_{\cdot j} ||_2 r \big| > 2\lambda_2 \eta_0 = 2\sqrt{\lambda_0 \lambda_2} +  \lambda_1  \implies j \in \mathrm{supp}(\bar{\beta}). 
\end{align*}
According to the derived dual objective~\eqref{eq:dual} and equations~\eqref{eq:B_frak}-\eqref{eq:Psi_1}, a feature's activity is determined by its product with the optimal dual variable $\bar{\alpha}$, e.g., $\eta_j(\bar{\alpha})$ for feature $j$. The dual range estimation $\bar{\alpha}$ ($B(\alpha;r) $)  allows us to perform feature screening in order to improve algorithm efficiency by following the approach for LASSO~\citep{Fercoq2015,GAP,Ndiaye2017}. As the support set $S$ is unknown, we just set $\gamma = \mu$ to ensure the ``safety'' of feature screening. Here ``safety'' means that the screening operation does not remove any feature belonging to $S=\mathrm{supp}(\bar{\beta})$. The framework proposed in this paper lays a broader bridge between screening methods and the solutions of $\ell_0$ regularized problems.

\section{Algorithm}\label{sec:algorithm}
With the strong duality regarding the Lagrangian form $L(\beta, \alpha)$~\eqref{eq:lagrangian}, we first develop a primal-dual algorithm to update both $\alpha$ and $\beta$. Then we propose a dynamic incremental method to reduce algorithm complexity. 
The dual objective $D(\alpha)$ is a non-smooth function as the term $\Psi()$ regarding $\alpha$ is non-smooth due to the truncation operation.  We focus on the following simplified dual form 
\begin{align} \label{eq:dual2}
   &\max_{\alpha \in \mathcal{F}^n} -  \sum_{i=1}^n  l_i^*(\alpha_i)   + \sum_{j=1}^p \Psi(\eta_j(\alpha); \lambda_0,  \lambda_1, \lambda_2),  \ \text{and}\\
 \label{eq:psi}
&\Psi(\eta_j(\alpha); \lambda_0,  \lambda_1, \lambda_2) 
 =   \begin{cases}
- \lambda_2 (|\eta_j(\alpha)| - \frac{\lambda_1}{2 \lambda_2})^2 + \lambda_0   &\text{if} \  \     |\eta_j(\alpha)|  \geq \eta_0 \\
0  &\text{if}  \  \   |\eta_j(\alpha)|  < \eta_0
\end{cases} .
\end{align}
The primal dual link is 
\begin{align} \label{eq:bigB}
&  \beta_j(\alpha)  =    \mathfrak{B}(\eta_j(\alpha)) = \begin{cases}
\mathrm{sign}\big(\eta_j(\alpha)\big) \big(|\eta_j(\alpha)| -\frac{\lambda_1 }{2\lambda_2} \big) &\text{if} \  \    |\eta_j(\alpha)|  \geq \eta_0 \\
0  &\text{if}  \  \   |\eta_j(\alpha)|  < \eta_0
\end{cases}
.
\end{align}
The super-gradient  regarding the dual variable  $\nabla_{\alpha} D(\alpha) =  X\beta(\alpha) -{l^*}^{'}(\alpha)$ can be taken as the partial derivative of $L(\beta, \alpha)$. We  give the dual problems of two objective functions in the supplements, and we will focus on linear regression to present the proposed algorithms. 

\vspace{-0.05in}
\subsection{Primal-dual Updating for Linear Regression}
\vspace{-0.05in}
 We use linear regression as an example to illustrate the proposed primal-dual inner solver of $\ell_0$ regularized problems.
For least square problem, the primal form is
\begin{align*}
&\min_{\beta \in \mathbb{R}^p} P(\beta) = f(\beta) + \lambda_0 ||\beta||_0, \ \text{and} \ f(\beta) =   \frac{1}{2}|| y - X\beta||^2_2   +  \lambda_1 ||\beta||_1 +\lambda_2 ||\beta||_2^2.
\end{align*}
For least square problem, $l_i(u; y_i) = \frac{1}{2}(y_i - u)^2$, then
$l_i^*(\alpha_i)=  \frac{1}{2}\big((y_i + \alpha_i)^2 - y_i^2 \big) =\frac{1}{2} \alpha_i^2 +y_i\alpha_i $. Here $u= x_i^{\top} \beta$. Thus the dual problem is
\begin{align} \label{eq:dual3}
\max_{\alpha}  D(\alpha) =   & - \frac{1}{2}\alpha^{\top}\alpha  - y^{\top}\alpha  +  \sum_{j=1}^p \Psi( - \frac{1}{2\lambda_2} \sum_{i=1}^n \bar{\alpha}_i x_i; \lambda_0, \lambda_1, \lambda_2).
\end{align}
The corresponding super-gradient can be easily computed, i.e.,
$g_{\alpha} =  \nabla_{\alpha} D(\alpha) = X\beta - Y - \alpha.$ After the super gradient ascent for the dual variables, we apply the primal-dual link function to get the variable in the primal space.

The dual objective $D(\alpha)$ is non-smooth. The super-gradient $g_{\alpha} =  \nabla_{\alpha} D(\alpha)$ can be improved with a more accurate primal variable estimation regarding the Lagrangian form~\eqref{eq:lagrangian}.  
We use coordinate descent~(CD)~\citep{Hazimeh18} to improve the estimation of  primal variable as
\begin{align} \label{eq:thresh}
&\beta_j = T(\mathbf{\beta}; \lambda_0, \lambda_1, \lambda_2)   =  \begin{cases}
\mathrm{sign}(\tilde{\mathbf{\beta}}_j) \frac{|\tilde{\mathbf{\beta}}_j| - \lambda_1}{1 + 2\lambda_2} \quad  &\text{if} \ \frac{|\tilde{\mathbf{\beta}}_j| - \lambda_1}{1 + 2\lambda_2}  \geq \sqrt{\frac{2\lambda_0}{1+2\lambda_2}} \\ 
0 \quad  &\text{if} \ \frac{|\tilde{\mathbf{\beta}}_j| - \lambda_1}{1 + 2\lambda_2}  < \sqrt{\frac{2\lambda_0}{1+2\lambda_2}} 
\end{cases} .
\end{align}

The proposed primal-dual updating procedure is given by Algorithm~\ref{alg:primal_dual_inner}. 
The primal coordinate descent $T()$ improves the solution from primal-dual relation $ \mathfrak{B}$. $\omega_t$ is the step size at $t$, and should be decreasing with $t$.  We use Algorithm~\ref{alg:primal_dual_inner} as the backbone solver in our primal-dual algorithm, and  $\overline{DGap}$ is the  sub-problem's duality gap    achieved by the inner solver.

\begin{wrapfigure}{R}{0.5\textwidth}\vspace{-5mm}
 \hspace{0.1in}
\begin{minipage}[t]{0.5\textwidth}
\begin{algorithm} [H]
\hrulefill\\
\vspace{-0.05in}
\caption{Inner solver with primal-dual  updating  }\label{alg:primal_dual_inner} \vspace{2mm}
	\KwInput{data $\{X, y\}$; $\lambda_0$, $\lambda_1$, $\lambda_2$; step size $\omega$; initial $\mathbf{\alpha}^0$,  $\mathbf{\beta}^0$}
	\KwResult{ $\mathbf{\alpha}^t$, $\mathbf{\beta}^t$ \vspace{-0.1in}}
	\hrulefill\\	
	 $t\leftarrow 0$;\\ 
	\While{$\overline{DGap}$ decreasing}{
	\textcolor{blue}{//Super-gradient } \\
	 $g_{\mathbf{\alpha}}^{t} =[{\beta^{t-1}}^{\top} x_1-  {l^{*}_1}^{'}(\alpha^{t-1}_1), ..., {\beta^{t-1}}^{\top} x_n - {l^{*}_n}^{'}(\alpha^{t-1}_n ) ]^{\top}$; \\
	  \textcolor{blue}{//Dual ascent with feasible projection } \\
	 $\alpha^t  =\mathcal{P}_{\mathcal{F}}( \alpha^{t-1} + \omega g_{\mathbf{\alpha}}^{t})$ ;  \\
	 $\eta(\alpha^t) = - \frac{1}{2\lambda_2} \sum_{i=1}^n \alpha^t_i x_i$; \\
	 \textcolor{blue}{//Primal-dual relation} \\
	 $ \mathbf{\beta}^t  =   \mathfrak{B}(\eta_j(\alpha^t); \lambda_0, \lambda_1, \lambda_2 ) $;  \\
	  \textcolor{blue}{//Primal coordinate descent} \\
	 $ \mathbf{\beta}^t = T(\beta^t; \lambda_0, \lambda_1, \lambda_2 ) $; \\
	 Compute duality gap $\overline{DGap}$ with $\beta_t$ and $\alpha_t$ ; \\
	 $t\leftarrow t+1$; 
	}
    \hrulefill\\
\end{algorithm}
\end{minipage}%
\vspace{-0.1in}
\end{wrapfigure}

Moreover, according to Remark~\ref{rmk:dual_condt}, the optimal dual satisfies $\bar{\alpha} \in (l^{*'})^{-1}(X\bar{\beta}) \cap \mathcal{F}^n$. For a   solver in primal space, we can use this equation to find a point in dual space and then compute the duality gap to evaluate the current solution. For linear regression, with $\beta$ we have $\alpha = X\beta -y $, then we can compute the duality gap via the  primal-dual gap  given by
\begin{align}\label{eq:dgap}
\xi(\beta, \alpha) = P(\beta) - D(\alpha).
\end{align}

Here $\tilde{\mathbf{\beta}}_j = (y - X\mathbf{\beta})^{\top} x_{\cdot j} + \mathbf{\beta}_j  x_{\cdot  j}^{\top}x_{\cdot j}$, and $x_{\cdot j}$ is the $j^{th}$ column of $X$, and it is also named the $j^{th}$  feature. The operation $T()$ always decreases the primal objective, i.e. with $\beta_b = T(\beta_a)$, we always have $P(\beta_b) \leq P(\beta_a)$, and hence a smaller duality gap.

\subsection{Improve Efficiency with Active Incremental Strategy}\label{sec:increment_strategy}

For sparse models, most of the features are redundant and they  incur extra computation costs.  The derived dual problem structure and the duality property provide an approach to implement feature screening~\citep{Fercoq2015,GAP,Ndiaye2017} and feature active incremental strategy~\citep{ren2017scalable,ren2020thunder}. According to the analysis in Section~\ref{sec:dual_v_est}, the activity of a feature $x_{\cdot j}$ depends on the value of $\eta_j(\bar{\alpha})$, i.e.,  $|\eta_j(\bar{\alpha})| = \frac{1}{2\lambda_2} |x^{\top}_{\cdot j} \bar{\alpha}|$. We use the current estimation range of $\bar{\alpha}$, i.e., $B(\alpha^s; r^s)$ to approximate the value of $\eta_j(\bar{\alpha})$. Here $s$ is the step number in the outer loop of the algorithm, and $\beta^s$ and $\alpha^s$ are the primal-dual solutions at step $s$. According to Theorem~\ref{Thm:ball}, the ball radius $\bar{\alpha}$ that depends on the duality gap at step $s$ is given by
\begin{align}\label{eq:rs}
r^s= \sqrt{ \frac{2(P(\beta^s) - D(\alpha^s)) } {\gamma}}.
\end{align}

\begin{algorithm}
\hrulefill \\
	 \caption{Dynamic Incremental  Algorithm }\label{alg:imprv_primal_dual}
\KwInput{ $X$, $y$; $\lambda_0$, $\lambda_1$, $\lambda_2$; $\xi$ }
\KwResult{ $\beta^s$ }
\hrulefill \\
     Choose a small set of features $\mathcal{A}_0$  from $X$ in the descending order of $|X^{\top}l'(\mathbf{0})|$, and $\mathcal{R}_0$ represents the rest features; \\ $s \leftarrow 0$; $\beta^0 \leftarrow 0$; $\alpha^0 \leftarrow 0$; $DoAdd = True$; \\
	\While{\textbf{True}}
	{
    	\textcolor{blue}{//Sub-problem solver} \\
	    $\alpha^s, \tilde{\beta}^s \leftarrow$ Solve the sub-problem with feature set $\mathcal{A}^s$   via Algorithm~\ref{alg:primal_dual_inner}; \\  
	    $\beta^s \leftarrow$ put $\tilde{\beta}^s$ in a size $p$ vector and set entries not in $\mathcal{A}^s$  zero;
	    
		\textcolor{blue}{//Dual range estimation} \\
		  Compute  the duality gap $DGap$ and  the ball region $B(\alpha^s; r^s)$ with~\eqref{eq:rs}; 
		  
		\If{$ DGap <\xi$  }
		{
		  \textbf{Stop}; \textcolor{blue}{//Algorithm exits }
		    }
		Feature Screening with $B(\alpha^s; r^s)$ and~\eqref{eq:screen_rule};  \\
			\If{ $DoAdd$ }
    		{\textcolor{blue}{//Feature Inclusion} \\
    		    \If{ $\max_{j \in \mathcal{R}^s} |x^{\top}_{\cdot j} \alpha^s| + \|x_{\cdot j} \|_2 r^s  <2\lambda_2 \eta_0$ }
        		{
        		    $DoAdd = False$; Continue;
        		}
    	    	Feature Inclusion with $\alpha^s$ and Algorithm~\ref{alg:add}, and update $\mathcal{A}^s$ and $\mathcal{R}^s$; 
    		 }
    	 $s\leftarrow s+1$;  
}
\hrulefill \\
\end{algorithm}

The proposed primal-dual algorithm for $\ell_0$ is given by Algorithm~\ref{alg:imprv_primal_dual}.  Algorithm~\ref{alg:imprv_primal_dual} starts with a small active set $\mathcal{A}$, then increases the active set's size after solving each sub-problem. We use $\mathcal{R}$ to represent the set of features not used by the sub-problem solver. The feature-inclusion algorithm is given by Algorithm~\ref{alg:add}. Moreover, we can derive a gap-screening algorithm~\citep{Fercoq2015,GAP} by using the upper bound of $|\eta_j(\bar{\alpha})|$'s approximation given in Section~\ref{sec:dual_v_est}.
Based on the derivation in Section~\ref{sec:dual_v_est}, we use the following safe principle for feature screening. 
\begin{align} \label{eq:screen_rule}
&\textbf{Feature Screening:}\ |x^{\top}_{\cdot j} \alpha^s| + ||x_{\cdot j} ||_2 r^s  < 2\lambda_2 \eta_0 \implies j \notin \mathrm{supp}(\bar{\beta}), \ \text{remove}. 
\end{align}
 Here $2\lambda_2 \eta_0 = 2\sqrt{\lambda_0 \lambda_2} +  \lambda_1$.  The screening rule is safe because it is derived based on concavity of the dual problem. Base on~\eqref{eq:screen_rule}, we  derive a stopping condition for feature inclusion. If all features in $\mathcal{R}$ satisfy~\eqref{eq:screen_rule}, we stop Feature Inclusion. This allows us to avoid redundant computation  resulting from some inactive features.
\begin{wrapfigure}{R}{0.5\textwidth}
\vspace{-0.2in}
\hspace{0.1in}
\begin{minipage}[t]{0.5\textwidth}
\begin{algorithm}[H]
\hrulefill \\
\caption{Feature Inclusion}\label{alg:add}
\KwInput{$\alpha^s$, $r^s$, $\mathcal{A}^s$, $\mathcal{R}^s$ } 
	\KwResult{ $\mathcal{R}^{s+1}$, $\mathcal{A}^{s+1}$  }
	\vspace{-0.1in}
	\hrulefill \\
	Set  $h = \lceil c\log(p) \rceil$, \, $\mathcal{H}\leftarrow$ Select $h$ features according to the descending order of $|x_{\cdot j}^{\top}\alpha^s|, j\in \mathcal{R}^s$;
	
	$\mathcal{A}^{s+1} \leftarrow \mathcal{A}^{s} \cup \mathcal{H}$, \  \ 
	$\mathcal{R}^{s+1} \leftarrow \mathcal{R}^{s} \setminus \mathcal{H}$ ; \\
	\hrulefill\\
\end{algorithm}
\end{minipage}%
\vspace{-0.1in}
\end{wrapfigure}

In Algorithm~\ref{alg:imprv_primal_dual}, the initialization values of $\beta^0$ and $\alpha^0$ are set to zero.  We use $DGap$ to represent the duality gap of the original problem attained by the primal-dual algorithm. We use $s$ as  iteration step indices in Algorithms~\ref{alg:add} and~\ref{alg:imprv_primal_dual}, to differentiate  from steps ($t$s) in Algorithm~\ref{alg:primal_dual_inner}.  Empirically, feature screening does improve algorithm efficiency. The feature active incremental strategy can significantly avoid redundant computation to achieve~the~target~duality~gap.

\vspace{-0.05in}
\section{Algorithm Analysis}\label{sec:analysis}
\vspace{-0.05in}
We discuss algorithm convergence in this section. The outer loop in Algorithm~\ref{alg:imprv_primal_dual} involves both feature screening and feature inclusion operations relying on the dual variable estimation for the original problem, i.e., $|| \alpha^s -  \bar{\alpha}||_2  \leq r^s, r^s = \sqrt{ \frac{2(P(\beta^s) - D(\alpha^s)) } {\gamma}}$ defined in Theorem~\ref{Thm:ball}.  Here $\gamma = \mu + \frac{\sigma_{min}(X_S)}{2 \lambda_2}$. As discussed in  Section~\ref{sec:dual_v_est}, we set  $\gamma = \mu$ to ensure the safety of feature screening. 

\begin{theorem}\label{Thm:outer_loop}
Assume that $l_i$ is $1/\mu$-smooth,  $||x_{i}|| \leq \vartheta,  \ \forall 1\leq i \leq n$, and $||x_{\cdot j}||= 1,  \ \forall 1\leq j \leq p$. Given a stopping threshold $\xi$, the complexity of Algorithm~\ref{alg:imprv_primal_dual} is  $O\bigg( 4n \bar{c}\overline{|\mathcal{A}|} |\mathcal{A}^*| \Omega\big(\bar{\xi}, \widehat{\xi}, \widehat{c}\big) + 3 np |\mathcal{A}^*| +    4n c^{\mathbf{m}} |\mathcal{A}_{\mathbf{m}}| \Omega\big(\xi^S, \breve{\xi}, c^{\mathbf{m}}\big) +  4n c^{*}  |\mathcal{A}^*|\Omega\big(\xi, \widetilde{\xi},  c^*\big) \bigg)$.  Here $\mathbf{m}$ is the number of outer steps to finish feature inclusion, and $\mathbf{m} \leq |\mathcal{A}^*| $. Moreover, $\overline{|\mathcal{A}|} = \sqrt{\frac{1}{\mathbf{m}}\sum_{s=1}^{\mathbf{m}}|\mathcal{A}_s|^2}$, $\bar{c}= \sqrt{\frac{1}{\mathbf{m}}\sum_{s=1}^{\mathbf{m}}(c^s)^2}$; $\widehat{\xi} = \max_{i:1 \leq i \leq \mathbf{m}} \xi^{s}_{s-1}$, $\widehat{c} = \max_{i:1 \leq i \leq \mathbf{m}} c^s$;   and $\breve{\xi} = \max \{\bar{\xi}, \xi^S\}$, $\widetilde{\xi} =\max \big\{\xi, \min \{\bar{\xi}, \xi^S\} \big\} $. 
\end{theorem}

\begin{remark}\label{rmk:out_loop}
The  screening operation~\eqref{eq:screen_rule} is safe, and it does not remove any features in $\mathcal{A}^s \cap \mathrm{supp}(\bar{\beta})$ at step $s$. With additional features added by the feature inclusion  operation (Algorithm~\ref{alg:add}), the primal objective $P(\beta^s)$  always decreases after  the solution of the sub-problem regarding feature set $\mathcal{A}^s$ using the inner solver (Algorithm~\ref{alg:primal_dual_inner}).
\end{remark}
The  screening operation usually  keeps the primal objective value $P(\beta^s)$ intact. With Remark~\ref{rmk:out_loop}, $\mathcal{A}^s$ and  $P(\beta^s)$  converge  after some steps, and also Algorithm~\ref{alg:imprv_primal_dual}  converges with  $DGap$ smaller than the given threshold $\xi$. In fact, the active incremental strategy  could significantly reduce redundant operations introduced by inactive features especially when the problem is with a high sparse level~\citep{ren2017scalable,ren2020thunder}. The solution sparse level (the size of $S=\mathrm{supp}(\bar{\beta})$) impacts the algorithm complexity. Additional analysis on algorithms  can be found in the supplemental file.

\vspace{-0.05in}
\section{Experiments}\label{sec:experiment}
\vspace{-0.05in}
Experiments focus on linear regression but our proposed algorithm can be extended to other forms of loss~functions. Via experimental studies, we show the effectiveness of our method by comparing with dual iterative hard thresholding~\citep{Liu17} and coordinate descent with spacer steps~(CDSS)~\citep{Hazimeh18} algorithms, which is the state-of-the-art solutions for the $\ell_0$ regularization problems. 

\vspace{-0.05in}
\subsection{Simulation Study}
\vspace{-0.05in}

In this study, we simulate the datasets under the linear regression setting, i.e., $\mathbf{y} = X\beta + \epsilon$. The data matrix is generated according to a multi-variate Gaussian $X_{n\times p}\sim \textrm{MVN}(0, \Sigma)$, and $\Sigma = (\sigma_{i,j})$. Exponential correlation~\citep{Hazimeh18} is utilized to control feature dependency  relationship, i.e., $\sigma_{ij} = \rho^{|i-j|}$ with $\rho = 0.4$. The noise $\epsilon$ is Gaussian white noise with $SNR=\{2, 5, 20\}$ . For the true parameters $\beta$, $3\%$ entries ($0.03p$) are randomly set to the values in $[-1.0, 1.0]$, and the rest ($0.97p$) are set to zero. We generate the datasets with $p=3\,000$  and $n$ varying 
in $\{200,  300, 400, 500, 600\}$. Each setting is replicated 50 times.

\begin{figure}[h]
 \centering
\mbox{\hspace{-0.25in}
\includegraphics[width=1.13in]{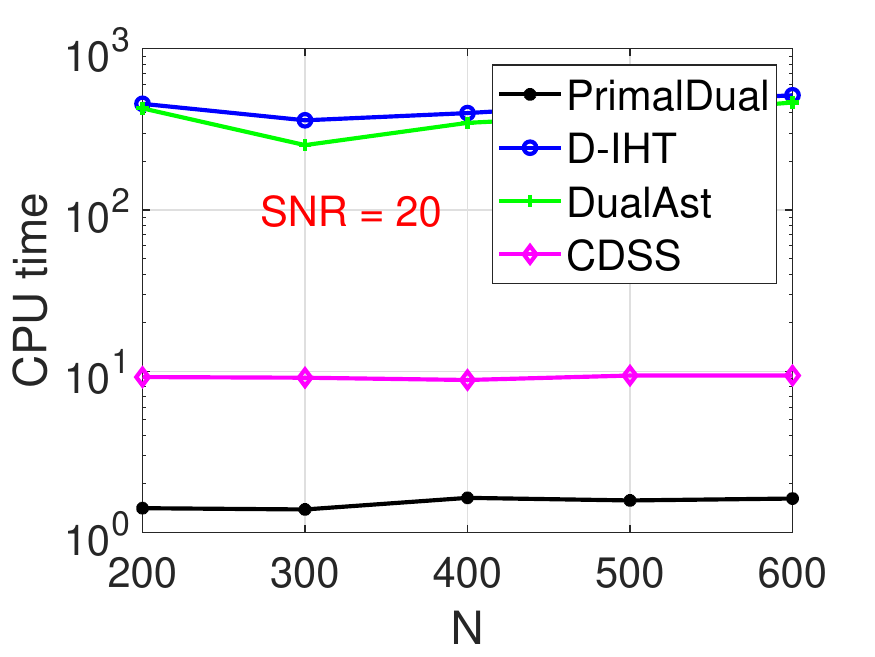}\hspace{-0.04in} 
\includegraphics[width=1.13in]{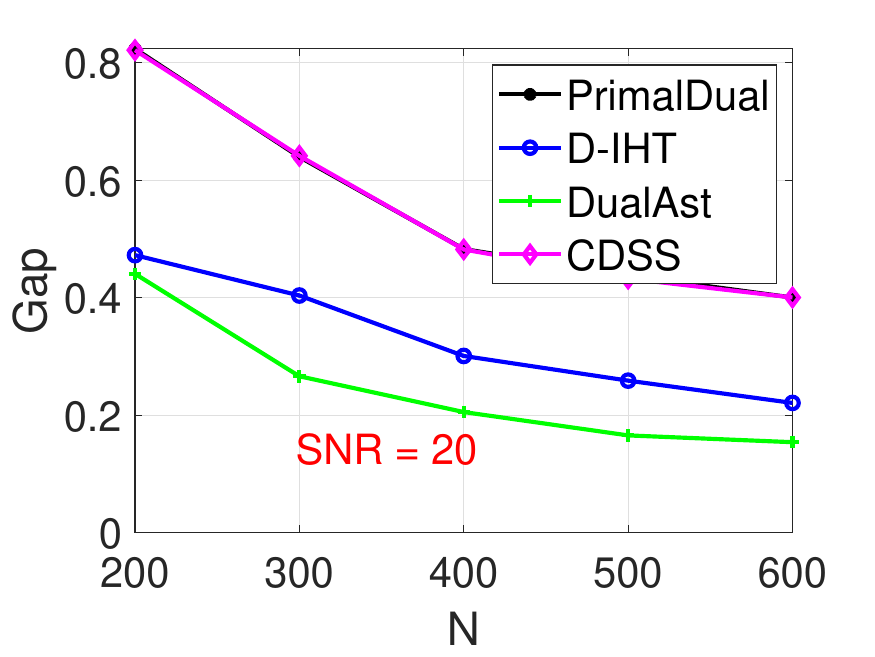}\hspace{-0.04in} \includegraphics[width=1.13in]{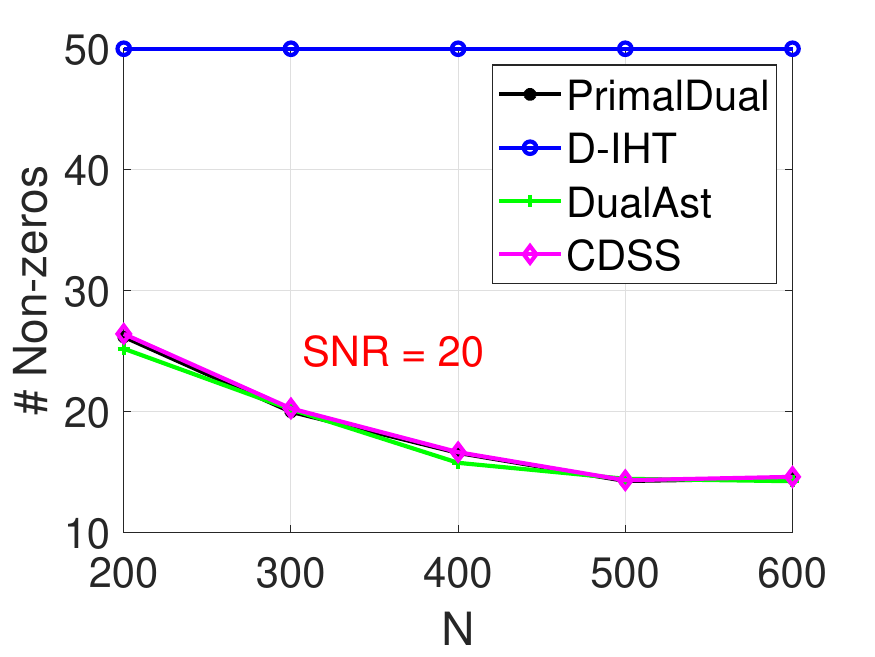} \hspace{-0.04in}
\includegraphics[width=1.13in]{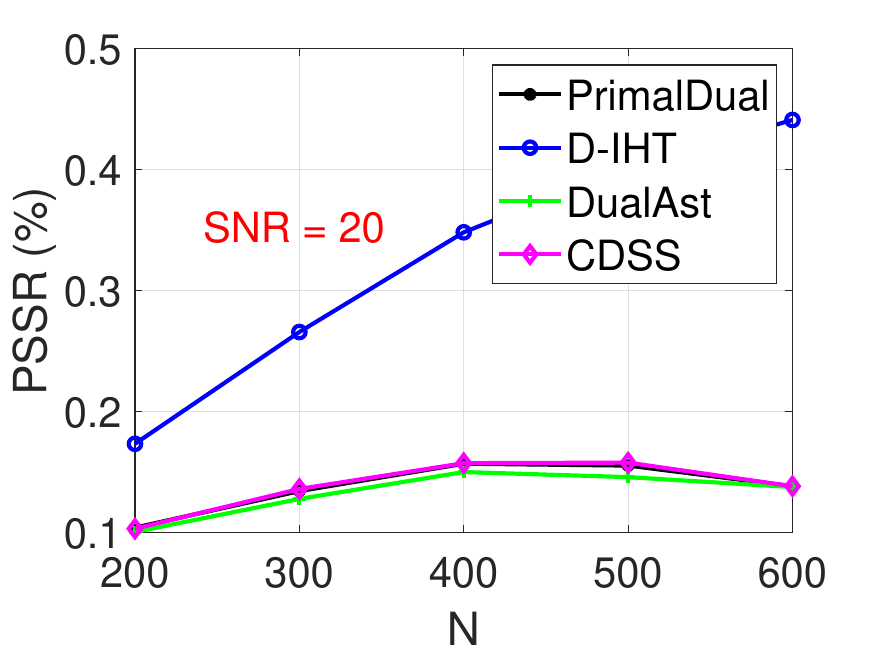} \hspace{-0.04in}
\includegraphics[width=1.13in]{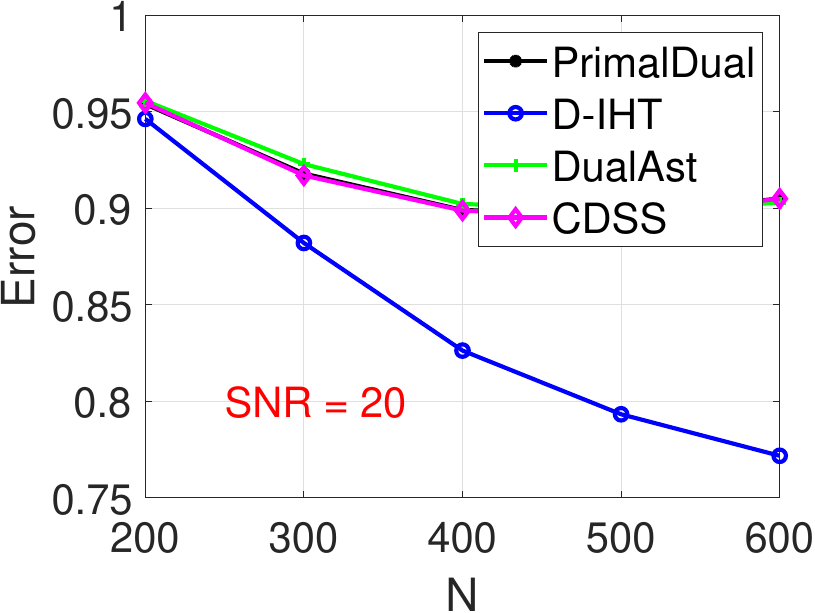}
}

\mbox{\hspace{-0.25in}
\includegraphics[width=1.13in]{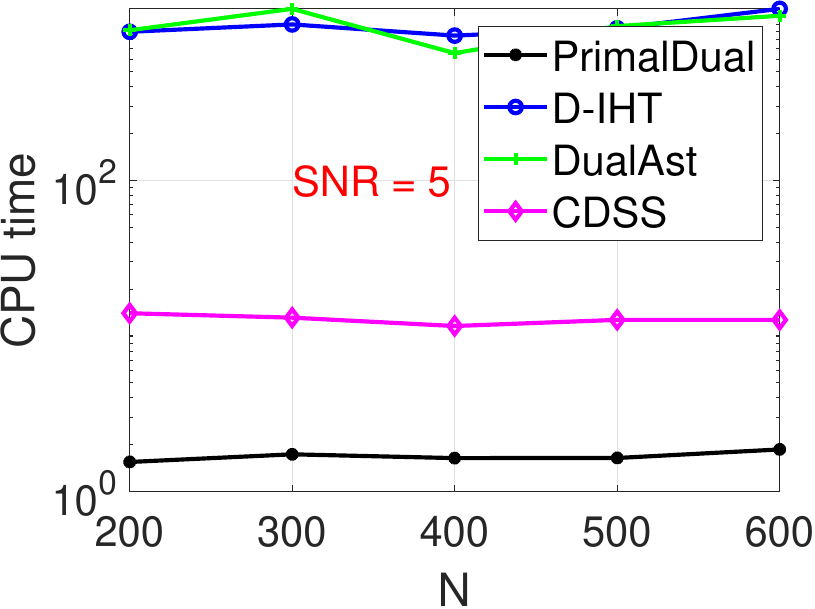}\hspace{-0.04in}
\includegraphics[width=1.13in]{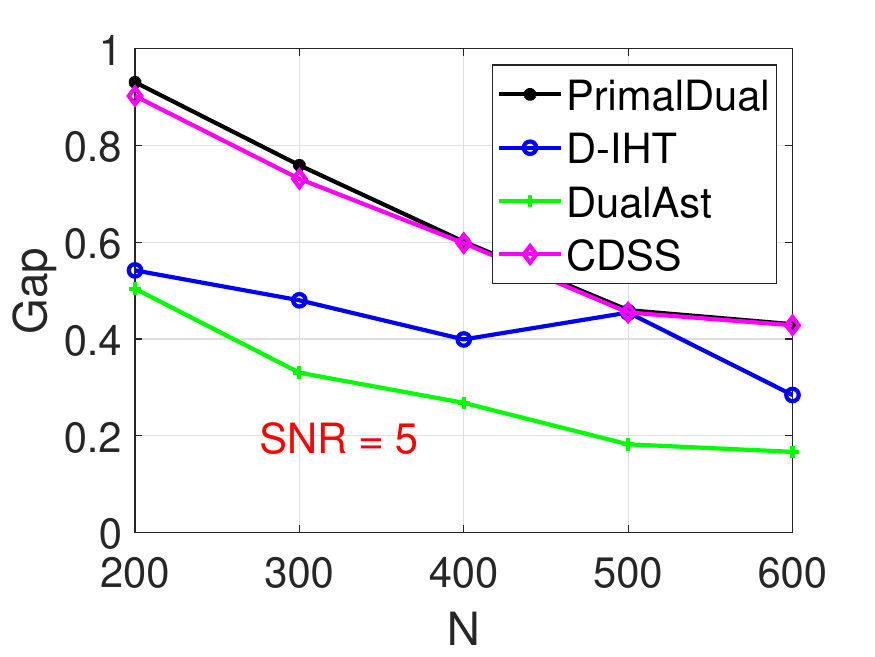} \hspace{-0.04in}
\includegraphics[width=1.13in]{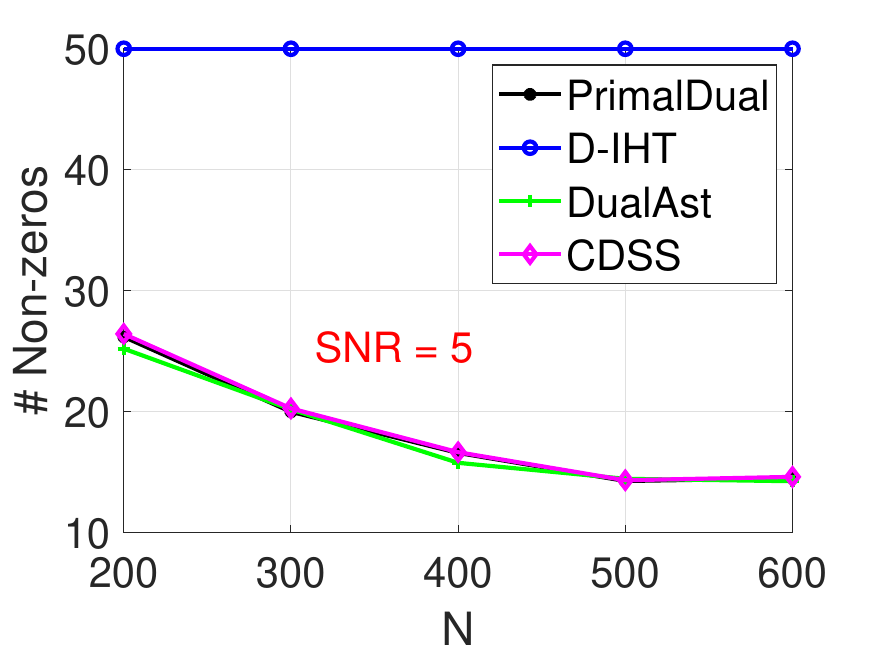} \hspace{-0.04in}
\includegraphics[width=1.13in]{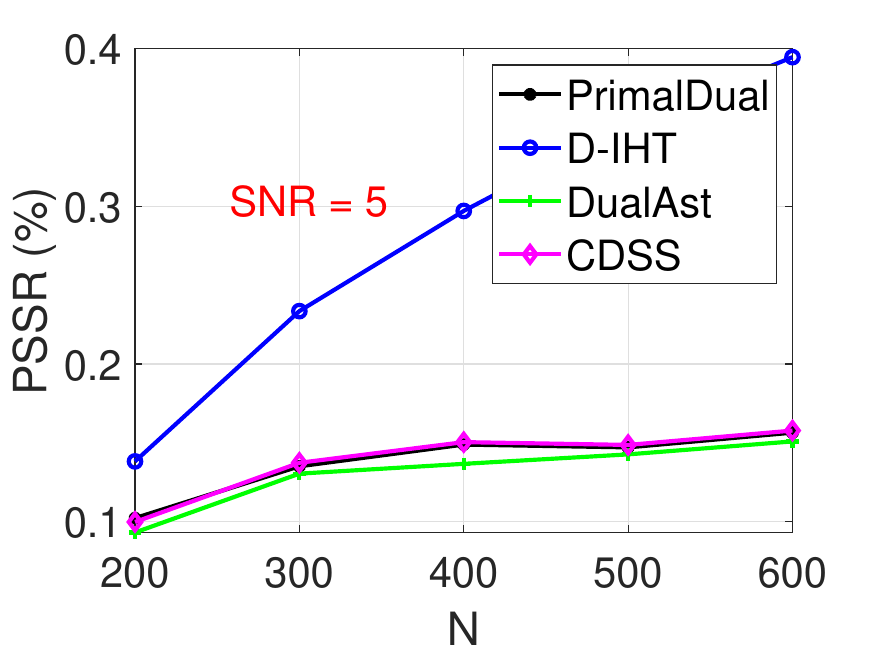} \hspace{-0.04in}
\includegraphics[width=1.11in]{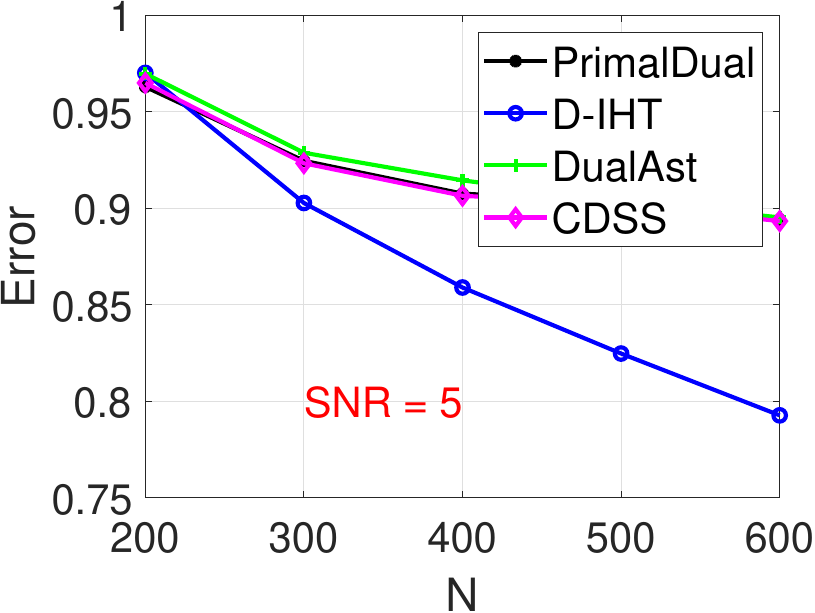}
}

\mbox{\hspace{-0.25in}
\includegraphics[width=1.13in]{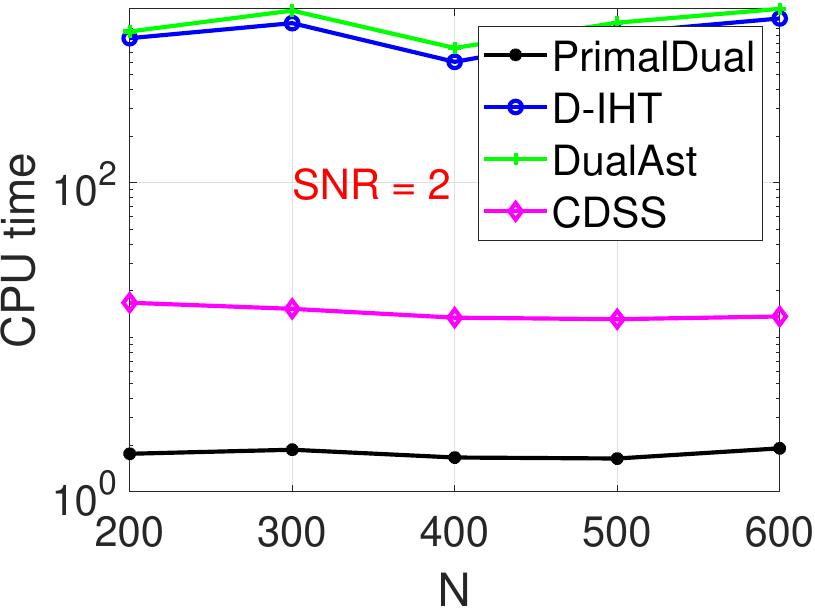}\hspace{-0.04in}
\includegraphics[width=1.13in]{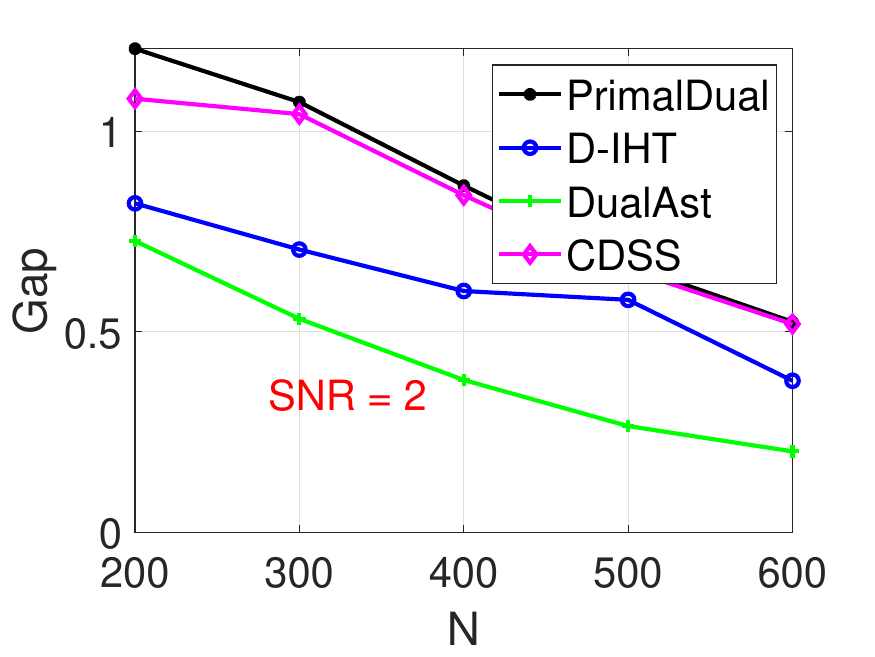} \hspace{-0.04in}
\includegraphics[width=1.13in]{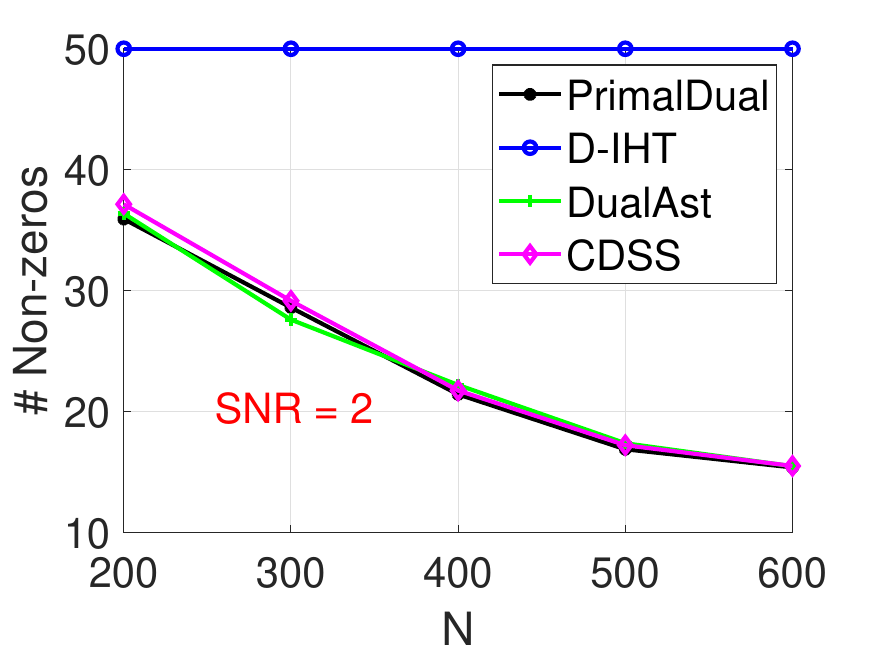} \hspace{-0.04in}
\includegraphics[width=1.13in]{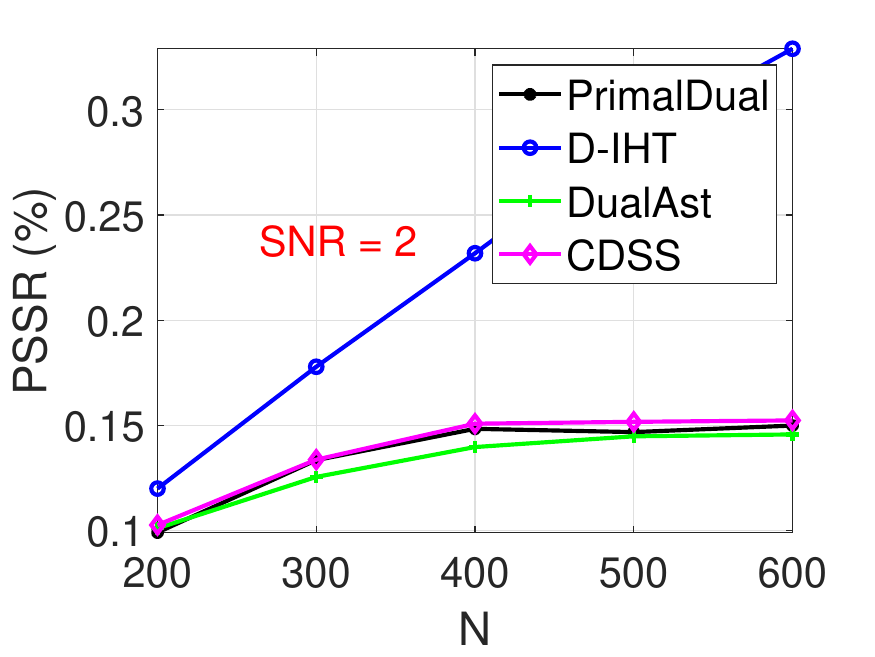} \hspace{-0.04in}
\includegraphics[width=1.14in]{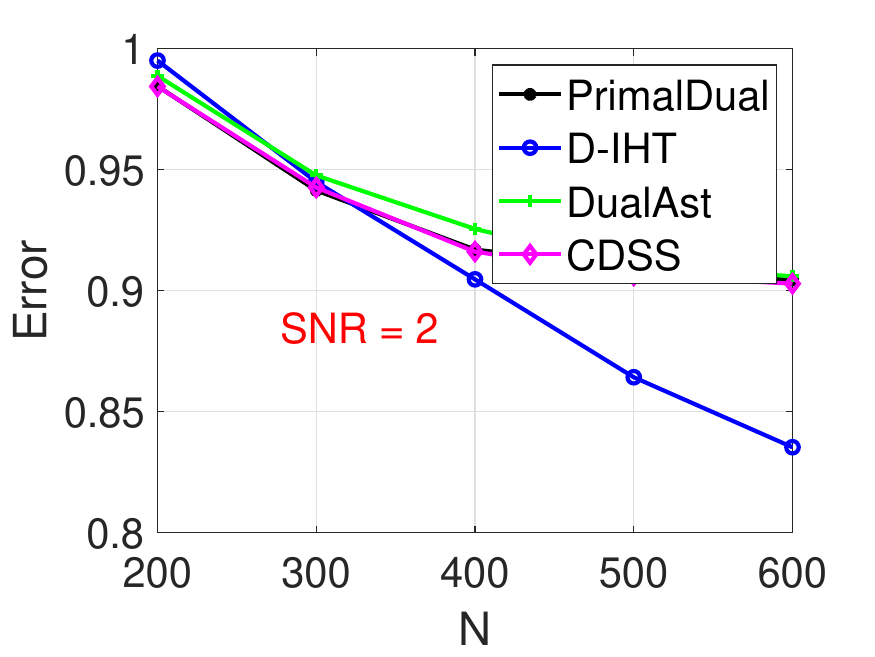}
}

\vspace{-0.1in}

\caption{Running time, duality gap, nonzero number,  {\it PSSR}, and estimation error for four algorithms on simulated data with $SNR \in \{20, 5, 2\}$. x-axis is the number of training samples,  from 200 to 600.}\label{fig:sim}
\vspace{-0.15in}
\end{figure}

We compare our primal-dual algorithm against the dual iterative hard thresholding~(Dual-IHT~\citep{Liu17}) and coordinate descent with spacer steps~\citep{Hazimeh18} algorithms. We use `D-IHT', and `CDSS' to represent the two algorithms, respectively. We use `PrimDual' to represent our proposed primal-dual algorithm, and `DualAst' to represent dual ascent method which is Algorithm~\ref{alg:primal_dual_inner} without the primal updating steps. 
All algorithms are implemented in Matlab and run in the same environment. 
D-IHT~\citep{Liu17} is a primal-dual method with dual ascending  using hard threshold to keep $k$ largest value of $|\beta|$ in the primal space. CDSS~\citep{Hazimeh18} is a coordinate descent method operates in the primal space enhanced with PSI~(Partial Swap Inescapable) as the stopping criteria. We use  the duality gap~($DGap$) threshold $\epsilon=1.0E-6$ as the the same stopping condition for D-IHT, DualAst and our PrimDual. 
The algorithms may require extremely long time to reach a small duality gap threshold.   We also use the duality change, i.e., $\zeta = |DGap^{t-2} - DGap^{t}|$ as a stopping condition for the three algorithms, and we set $\zeta = 1.0E-6$ in the experiments.  Moreover, we use the same learning rate $\omega= 0.000\,5$ for the three algorithms.

Two indices are adopted for evaluating the performance. The first one is the  {\it percentage of successful support recovery~(PSSR)}.  The second is parameter estimation error $|| \beta - \bar{\beta}||/||\bar{\beta}||$. Here $\bar{\beta}$ is the ground truth used in simulation. Figure~\ref{fig:sim} gives the performance of these four algorithms on datasets with different SNR values.
To achieve meaningful comparison, we choose $\lambda_0, \lambda_1$ and $\lambda_2$ to recover support number close  to the ground truth value. For the dataset with $SNR=20$, we use $\lambda_0 = 0.03, \lambda_1 = 0.02, \lambda_2 = 1.0$, and we set $\lambda_0 = 0.1, \lambda_1 = 0.2, \lambda_2 = 1.0$ for datasets with $SNR=2, 5$.
 From the plots, we can see that the proposed primal-dual algorithm can achieve similar {\it PSSR} and estimation error values (except for D-IHT since its sparsity is pre-determined), but use much less time. It shows that the proposed primal-dual algorithm and incremental strategy significantly reduce the redundant operations resulted from inactive features. 

\vspace{-0.05in}
\subsection{News20 Dataset}
\vspace{-0.05in}

After pre-processing, the commonly used News20 dataset contains 20 classes, $15\,935$ samples, and $62\,061$ features in the training set. The 20 labels in News20 dataset are transformed to response values ranging  $[-10,10]$ in the experiments. We randomly form five datasets with the number of samples ranging in $\{200, 300, 400, 500, 600\}$. We set the learning rate $\omega=0.0005$ for D-IHT, DualAst, and our PrimDual. The stopping conditions are $\epsilon=1.0E-6$ and $\zeta = 1.0E-6$. The hyper-parameters are set with $\lambda_0=0.1, \lambda_1=0.15, \lambda_1=1.0$. We set $c=4.0$ for Algorithm~\ref{alg:add} in our experiments.

\begin{figure}[h]
\vspace{-0.1in}
 \centering
\mbox{
\includegraphics[width=1.35in]{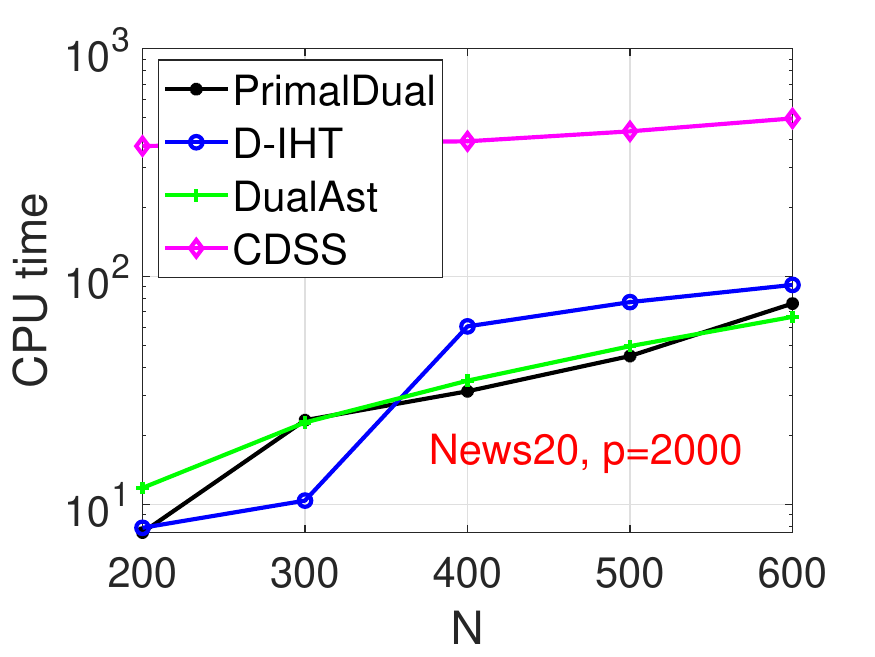}
\includegraphics[width=1.35in]{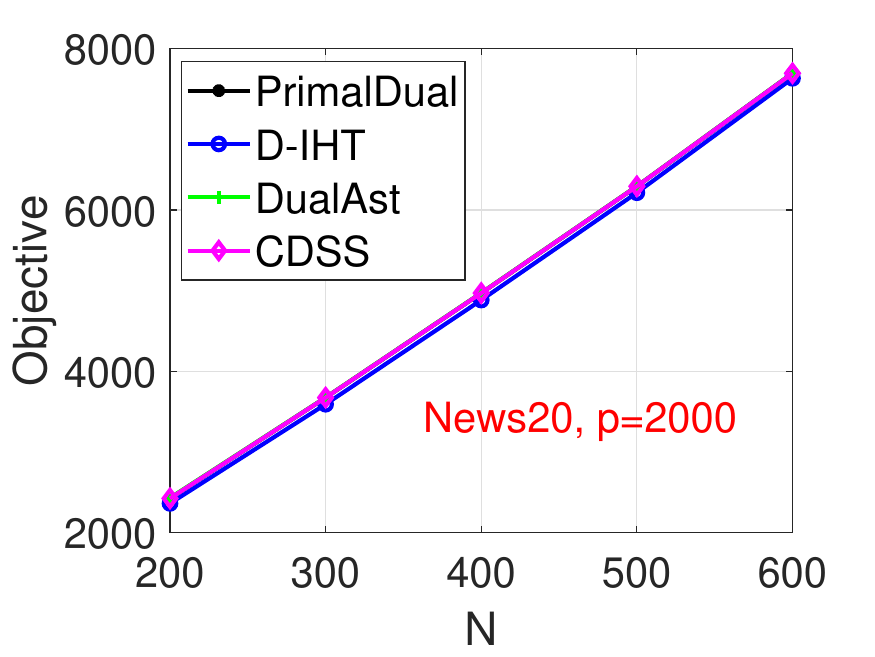}
\includegraphics[width=1.35in]{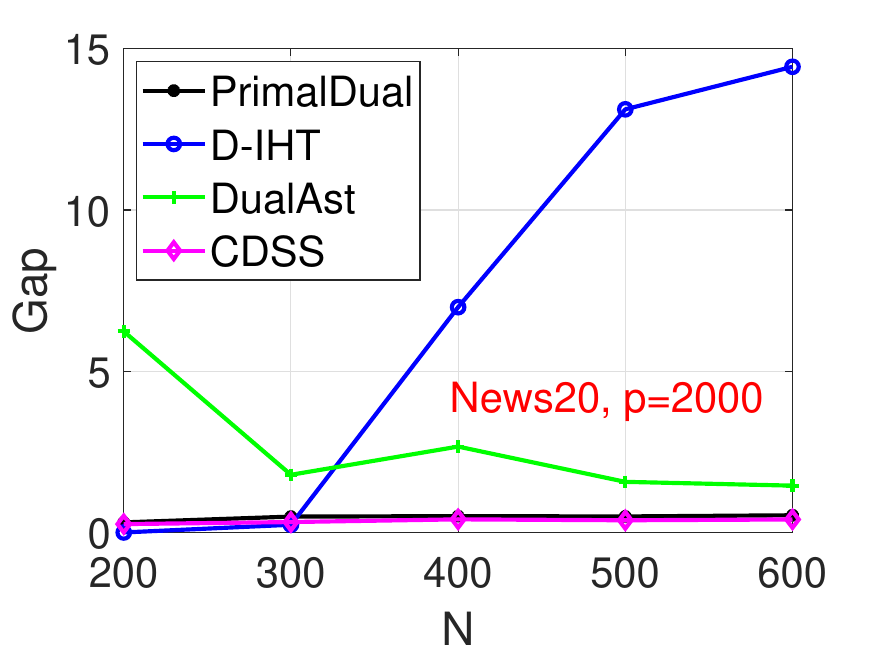}
\includegraphics[width=1.35in]{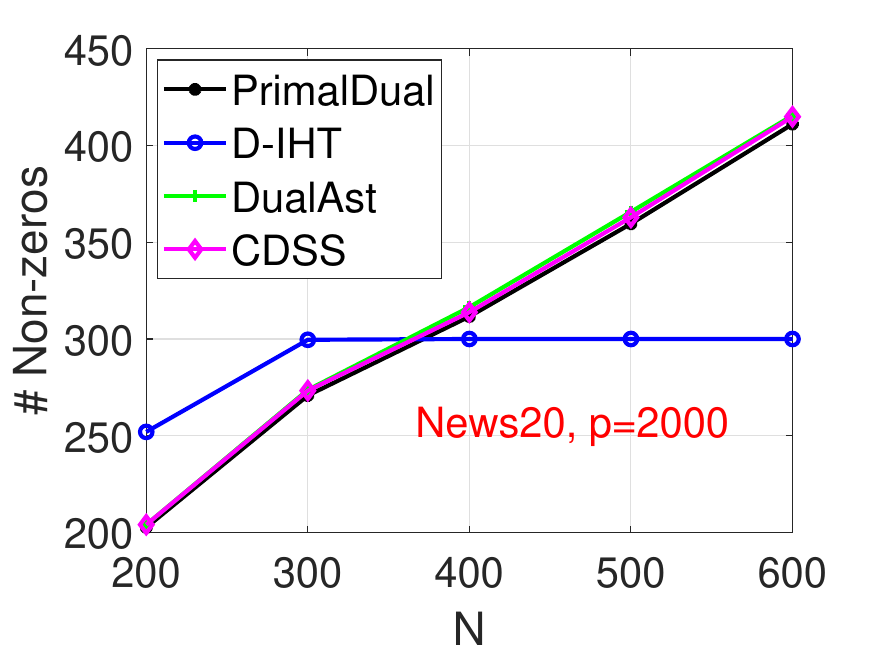}

}

\mbox{
\includegraphics[width=1.35in]{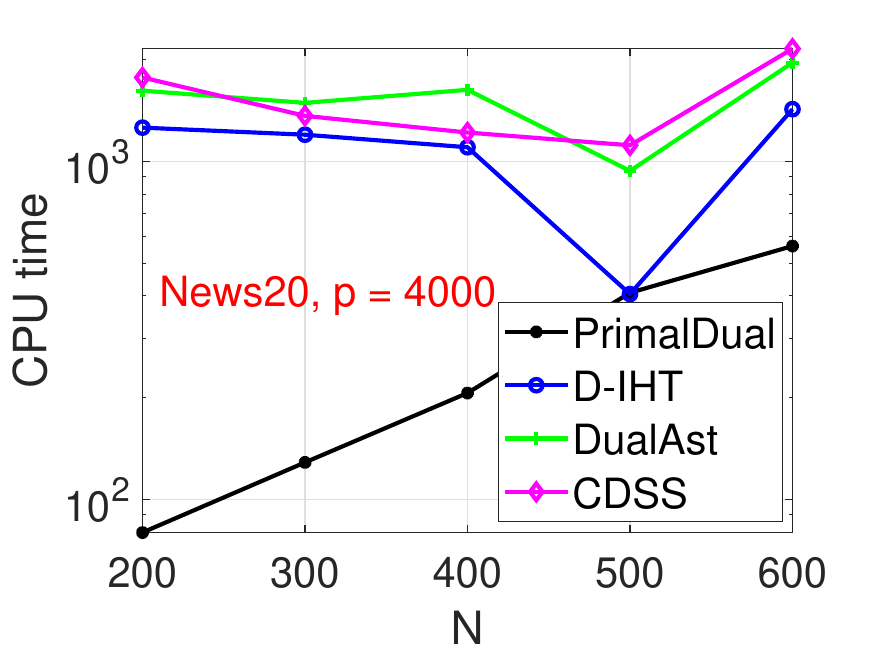}
\includegraphics[width=1.35in]{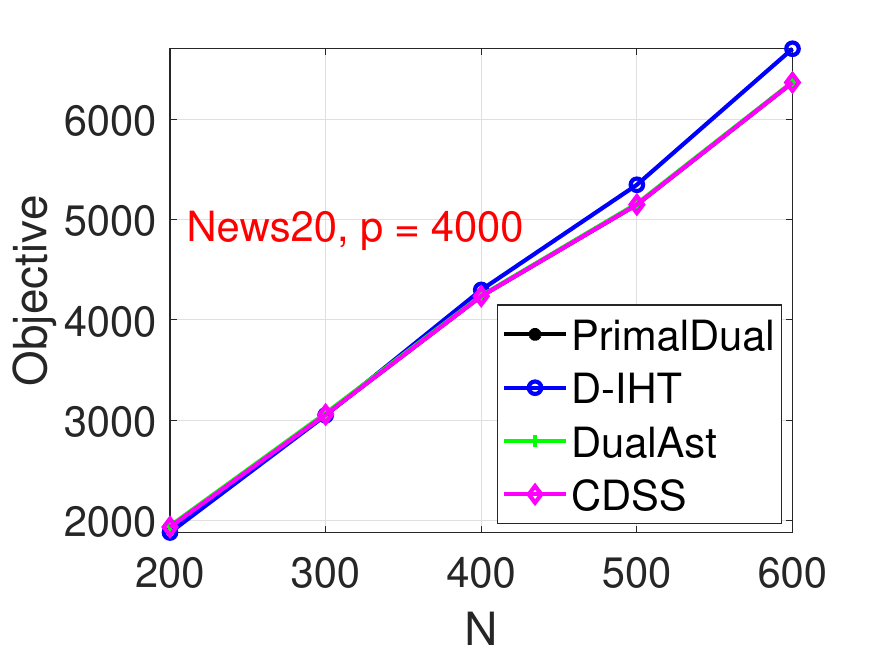}
\includegraphics[width=1.35in]{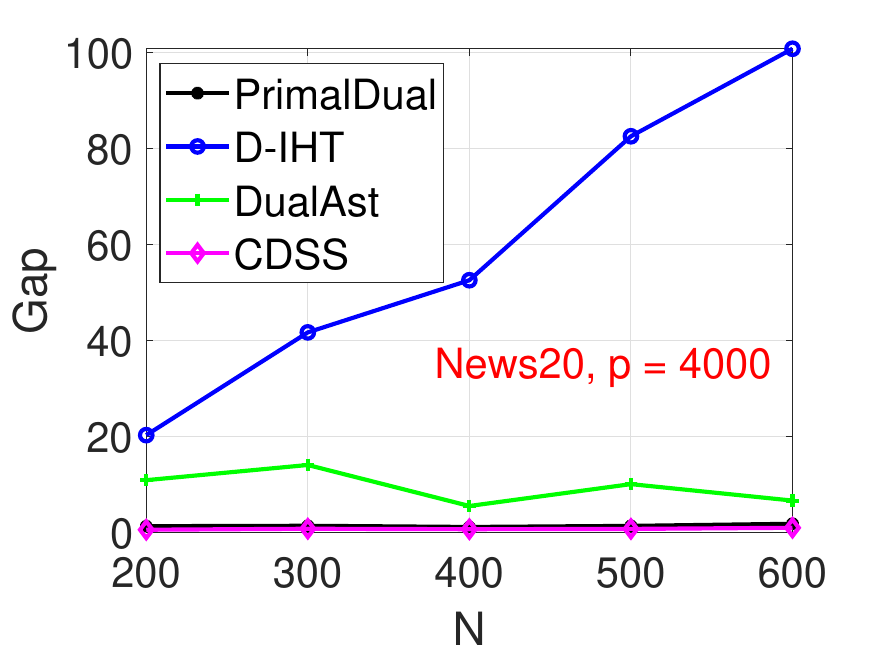}
\includegraphics[width=1.35in]{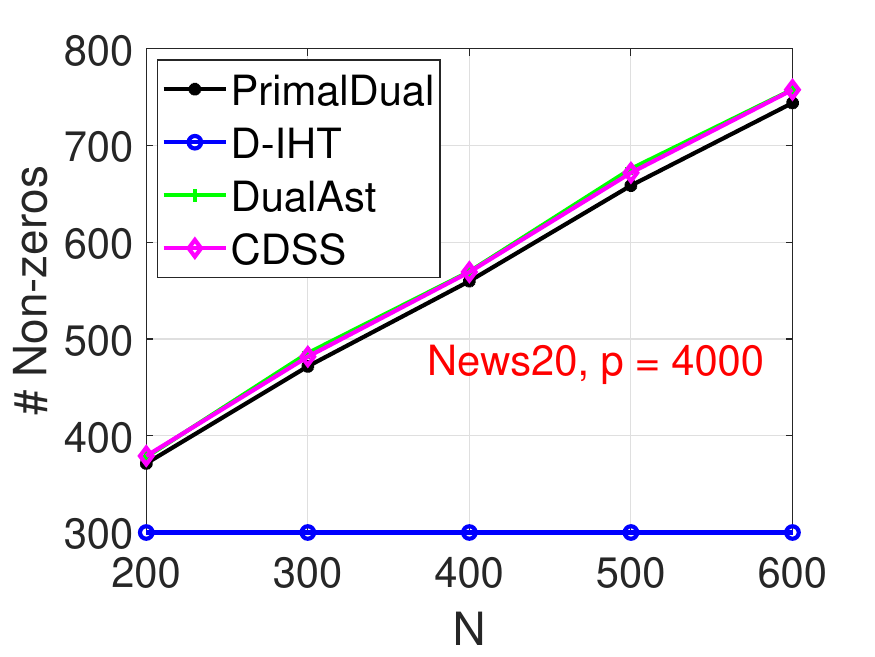}
}

\vspace{-0.1in}
\caption{The  running time, primal objective $P(\hat{\beta})$,  duality gap, and nonzero number  for different methods on News20 dataset. }\label{fig:news20}
\vspace{-0.1in}
\end{figure}

The upper row of Figure~\ref{fig:news20} shows the results of different methods on News20 with randomly select $p=2\,000$ features.
Each setting is replicated for 20 times.
We can see that under approximately the same primal objective and duality gap values, our  primal-dual method uses less computation time compared against other methods when $N$ becomes larger. Though DualAst has similar computation cost as our method, it cannot achieve small duality gap values on all cases. We notice that CDSS takes the longest time in this case, and it could be due to that the PSI stopping condition is hard to satisfy on some real-world datasets.

The bottom row of Figure~\ref{fig:news20} gives other results on News20 with a larger feature number (randomly selected $p=4\,000$ features). Hyper-parameters are set with $\lambda_0=0.10, \lambda_1=0.15, \lambda_2=1.00$. We set $k=300$ for D-IHT.  Each case is replicated for 10 times and the average result is reported.
From the plots, we can see that with longer running time, CDSS achieves the smallest duality gap values. However, our primal-dual algorithm takes much less computation cost to achieve similar solutions.  

Additional real-world experimental results can be found in the supplemental file.

\vspace{-0.05in}
\section{Discussion}\label{sec:discussion}
\vspace{-0.05in}

We provide more details on the differences between our work and related methods in this section to highlight our technical contributions. 

Our developed primal-dual method is different from the D-IHT method~\citep{Liu17,yuan2020dual}. First, our primal-dual method focuses on a different problem~\eqref{eq:primal1} from the one that the D-IHT solves~\eqref{eq:hard_k}. Apart from using soft regularization rather than hard constraint,~\eqref{eq:primal1} also includes the $\ell_1$ penalty that could be helpful in cases with low SNR values. Second, our primal-dual method performs updating in both primal and dual spaces to approach the saddle points, and it can potentially attain solutions with smaller duality gaps. Finally, the most important of all, our objective~\eqref{eq:primal1} and the derived dual form~\eqref{eq:dual}  allow us to employ screening and coordinate incremental strategies~\citep{Fercoq2015,GAP,ren2017scalable,ren2020thunder,atamturk2020safe} to significantly boost the efficiency of the algorithm.

There are several obvious differences between our primal-dual method and the coordinate descent with spacer steps~(CDSS)~\citep{Hazimeh18}. Different from our primal-dual method, CDSS utilizes coordinate descent in the primal space for parameter updating. In CDSS~\citep{Hazimeh18}, they also rely on partial swap inescapable (PSI-$\kappa$) to improve the solution with $\kappa=1$. PSI with $\kappa \geq 2$ will introduce much more extra computation that is usually not affordable.  Our primal-dual method employs coordinate incremental strategy to save computation cost.  Experimental results indicate that our proposed method can achieve similar solution quality as CDSS but with much less computation time. 

In~\cite{atamturk2020safe}, the authors proposed a screening method for $\ell_0$ regularized problems. However, their objective does not include the $\ell_1$ norm. Our proposed method focuses on a more generalized problem that is potentially more powerful on datasets with low SNRs. Moreover, besides the safe screening rule in Section~\ref{sec:dual_v_est}, the proposed coordinate incremental strategy introduced in Section~\ref{sec:increment_strategy} is empirically effective on different datasets. The screening  methods~\citep{Fercoq2015,GAP,Ndiaye2017} and coordinate incremental strategies~\citep{ren2020thunder,ren2017scalable} used for $\ell_1$ regularized problems can be taken as special cases of the proposed method with $\lambda_0=0$. 

This paper primarily focuses on the theoretical perspective of our proposed methodology. We have evaluated our primal-dual method on  real-world datasets with small data sample sizes ($n\leq 600$) and feature dimension ($p\leq 4000$). More real-world datasets and improved implementations will be tested with more general settings. 

\vspace{-0.05in}
\section{Conclusion}\label{sec:conclusion}
\vspace{-0.05in}
In this paper, we have studied the dual forms of a broad family of $\ell_0$ regularized problems. Based on the derived dual form, a primal-dual algorithm acren2017scalableated with active coordinate selection has been developed. Our theoretical results show that the reformed best subset selection problem can be solved with polynomial complexity. The developed framework and theory can be integrated with many feature screening strategies. 
Experimental results have shown that our primal-dual method can reduce redundant operations introduced by inactive features and hence reduce computational costs.

\bibliography{ref}
\bibliographystyle{plain}

\newpage\clearpage
\appendix

In this Appendix, we provide theoretical proofs. Additional remarks are given in Section~\ref{sec:add_remarks}. Section~\ref{sec:outloop_complex} delivers the outer loop convergence and complexity analysis of Algorithm~\ref{alg:imprv_primal_dual}; Section~\ref{sec:proof_thm} shows the proof of Theorem~\ref{Thm:ball}; Section~\ref{sec:supp_analysis} gives the inner algorithm analysis; Section~\ref{sec:proof_duality} delivers the derivation of the duality.

\
\

\section{Additional Remarks}~\label{sec:add_remarks} 

\begin{itemize}
    \item[] \textbf{Technical Contribution:}  This paper investigates the duality of the generalized sparse learning problem~\eqref{eq:primal1},  following~\cite{pilanci2015sparse,Liu17,yuan2020dual}. The  generalized form helps overcome over-fitting issues of $\ell_0$ regularized problems when the SNRs of given data are low~\citep{pmlr-v65-david17a,mazumder2022subset}. The proposed framework considers  active coordinate incremental and screening strategies~\citep{Fercoq2015,GAP,Ndiaye2017,atamturk2020safe,ren2017scalable,ren2020thunder}  by leveraging  the duality structure properties of the problem~\eqref{eq:primal1}. The quality of solutions can be evaluated by the duality gap~\eqref{eq:dgap} with the current dual solution $\alpha$ calculated through Remark~\ref{rmk:dual_condt}. 

     \item[] \textbf{Convergence Analysis:} Algorithm  convergence and complexity are studied for both inner and outer updating procedures. The concaveness of the dual problem and the strong duality ensures the attainability of the optimal solutions in polynomial computation complexity. Our analysis (Theorem~\ref{Thm:outer_loop} and Remark~\ref{rmk:outer_loop}) shows that the complexity of the proposed algorithm is proportional to the size of the optimal active set $\mathcal{A}^*$.

\end{itemize}

\newpage

\section{Outer Loop Analysis}~\label{sec:outloop_complex}
We present the complexity analysis of the outer loop convergence in this section. We first prove an equation that will be used in the following analysis. 

\begin{lemma}~\label{} 
Let $a >b>0$ and $\Delta = a-b$, then $a\log a -b \log b =  \Delta \log b  + a  \log \frac{a}{b}$.
\end{lemma} 

\begin{proof}
\begin{align*}
a\log a -b \log b &= (b+ \Delta ) \log (b+ \Delta ) - b \log b \\
&=  \Delta \log b  + a  \log (1+\frac{ \Delta}{b})\\
&= \Delta \log b  + a  \log \frac{a}{b}.
\end{align*}

\end{proof}

To make the presentation concise, we further define the following function: 
\begin{align*}
\Omega(\xi_1, \xi_2, c)=(\frac{1}{\xi_1}-\frac{1}{\xi_2})\log \frac{c}{\xi_2} + \frac{1}{\xi_1} \log \frac{\xi_2}{\xi_1}, \ \  \textrm{where } \xi_1, \xi_2, c>0.
\end{align*}

Algorithm~\ref{alg:imprv_primal_dual} generally has three stages: feature inclusion, feature screening, and accuracy pursuit. If all the active features are added to the active set $\mathcal{A}$  but the stopping threshold $\xi$ has not been reached, the label $DoAdd$ becomes False, and the algorithm enters the feature screening and accuracy pursuit stages.

Let $\mathcal{A}^*$ be the optimal active set of the original problem.  In the outer step $s$, we use 
$P^s()$ and $D^s()$ to denote the primal and dual objective values. Given a pair of $(\alpha, \beta)$, the duality gap regarding the sub-problem (with active set $\mathcal{A}_s$)  in step $s$  is given by  $\xi^s(\beta, \alpha) = P^s(\beta ) - D^s(\alpha)$. In addition, $ \upsilon_s$ is the computational complexity for one updating iteration in the inner Algorithm~\ref{alg:primal_dual_inner} in outer step $s$. 

Moreover, let $X_S^s$ be the feature matrix of sub-problem $s$, and here
$c^s = c_0\bigg(1+ \frac{\sigma_{max}(X^s_S)}{2\mu\lambda_2} \bigg)$. For the sub-problem with active set $\mathcal{A}^*$, we have $c^* = c_0\bigg(1+ \frac{\sigma_{max}(X_{\mathcal{A}^*})}{2\mu\lambda_2} \bigg)$.  Let $\bar{\xi}$ be the duality gap value that adds at least one feature to  the active set in each outer step;  $\xi^{s}_{s-1}$ is the initial duality gap value of the inner Algorithm~\ref{alg:primal_dual_inner} in outer step $s$;  $\xi^S$ is the duality gap  threshold of  feature screening, i.e., with duality gap $\xi < \xi^S $ all inactive features will be removed from $\mathcal{A}$ with operation~\eqref{eq:screen_rule}. The following theorem gives the complexity of Algorithm~\ref{alg:imprv_primal_dual}.

\begin{theorem}\label{Thm:outer_loop}
Assume that $l_i$ is $1/\mu$-smooth,  $||x_{i}|| \leq \vartheta,  \ \forall 1\leq i \leq n$, and $||x_{\cdot j}||= 1,  \ \forall 1\leq j \leq p$. Given a stopping threshold $\xi$, the complexity of Algorithm~\ref{alg:imprv_primal_dual} is  $O\bigg( 4n \bar{c}\overline{|\mathcal{A}|} |\mathcal{A}^*| \Omega\big(\bar{\xi}, \widehat{\xi}, \widehat{c}\big) + 3 np |\mathcal{A}^*| +    4n c^{\mathbf{m}} |\mathcal{A}_{\mathbf{m}}| \Omega\big(\xi^S, \breve{\xi}, c^{\mathbf{m}}\big) +  4n c^{*}  |\mathcal{A}^*|\Omega\big(\xi, \widetilde{\xi},  c^*\big) \bigg)$.  Here $\mathbf{m}$ is the number of outer steps to finish feature inclusion, and $\mathbf{m} \leq |\mathcal{A}^*| $. Moreover, $\overline{|\mathcal{A}|} = \sqrt{\frac{1}{\mathbf{m}}\sum_{s=1}^{\mathbf{m}}|\mathcal{A}_s|^2}$, $\bar{c}= \sqrt{\frac{1}{\mathbf{m}}\sum_{s=1}^{\mathbf{m}}(c^s)^2}$; $\widehat{\xi} = \max_{i:1 \leq i \leq \mathbf{m}} \xi^{s}_{s-1}$, $\widehat{c} = \max_{i:1 \leq i \leq \mathbf{m}} c^s$;   and $\breve{\xi} = \max \{\bar{\xi}, \xi^S\}$, $\widetilde{\xi} =\max \big\{\xi, \min \{\bar{\xi}, \xi^S\} \big\} $. 
\end{theorem}

\begin{proof}
In step $s$,  $h$ features are added to $\mathcal{A}_s$. Let's assume the ratio of active features in the adding operation to be  $r_s$, and $0 < r_s \leq 1$. Meanwhile, in Algorithm~\ref{alg:primal_dual_inner} sufficient updating iterations  are conducted to  ensure the duality gap $\xi^s$ is small enough ($\xi^s <= \bar{\xi}$ ) so that at least one feature is added, i.e., $hr_s \geq 1$.

 In the outer step $s$, the computing complexity for each updating iteration in the sub-problem~\ref{alg:primal_dual_inner} is $\upsilon_s$.  Let $T_s$ be the total iteration number of Algorithm~\ref{alg:primal_dual_inner} to reach $\xi^s$  for the sub-problem in outer step $s$. The complexity of each outer step includes computing the primal and dual objectives, feature including, and screening operations.  It is easy to see that the computing complexity of each outer step is $\varsigma np$, and here $0<\varsigma<3$. 

For the feature inclusion stage, the complexity is 
\begin{align*}
T_{add} =& \sum_{s=1}^{\mathbf{m}} \bigg( \upsilon_s T_s + \varsigma np \bigg) \\
=&\sum_{s=1}^{\mathbf{m}} \upsilon_s T_s  + \varsigma np \mathbf{m} \\
=& T_{a1}  + \varsigma np \mathbf{m} .
\end{align*}

The inner iteration number in outer step $s$ is $T_s=\frac{c^s}{\bar{\xi}} \log \frac{c^s}{\bar{\xi}} -  \frac{c^s}{\xi^{s}_{s-1}} \log \frac{c^s}{\xi^{s}_{s-1}}$.  
Let $\widehat{\xi} = \max_{i:1 \leq i \leq \mathbf{m}} \xi^{s}_{s-1}$ and $\widehat{c} = \max_{i:1 \leq i \leq \mathbf{m}} c^s$, with the equation in Lemma~\ref{Lem:support},  we get
\begin{align*}
T_{a1}=& \sum_{s=1}^{\mathbf{m}} \upsilon_s T_s \\
=&  \sum_{s=1}^{\mathbf{m}} \upsilon_s \bigg(\frac{c^s}{\bar{\xi}} \log \frac{c^s}{\bar{\xi}} -  \frac{c^s}{\xi^{s}_{s-1}} \log \frac{c^s}{\xi^{s}_{s-1}} \bigg)  \\
\leq &  \sum_{s=1}^{\mathbf{m}} \upsilon_s \bigg(\frac{c^s}{\bar{\xi}} \log \frac{c^s}{\bar{\xi}} -  \frac{c^s}{\widehat{\xi}} \log \frac{c^s}{\widehat{\xi}} \bigg) \\
=&  \sum_{s=1}^{\mathbf{m}} \upsilon_s \bigg( \big(\frac{c^s}{\bar{\xi}} - \frac{c^s}{\widehat{\xi}} \big) \log \frac{c^s}{\widehat{\xi}} + \frac{c^s}{\bar{\xi}} \log \frac{\widehat{\xi}}{\bar{\xi} }  \bigg) \\
=&\sum_{s=1}^{\mathbf{m}} \upsilon_s c^s \bigg( \big(\frac{1}{\bar{\xi}} - \frac{1}{\widehat{\xi}} \big) \log \frac{c^s}{\widehat{\xi}} + \frac{1}{\bar{\xi}} \log \frac{\widehat{\xi}}{\bar{\xi} }  \bigg) \\
\leq & \bigg( \big(\frac{1}{\bar{\xi}} - \frac{1}{\widehat{\xi}} \big) \log \frac{\widehat{c}}{\widehat{\xi}} + \frac{1}{\bar{\xi}} \log \frac{\widehat{\xi}}{\bar{\xi} }  \bigg) \sum_{s=1}^{\mathbf{m}} \upsilon_s c^s \\
\leq &\sqrt{\sum_{s=1}^{\mathbf{m}} \upsilon_s^2 } \sqrt{\sum_{s=1}^{\mathbf{m}} (c^s)^2 }  \bigg( \big(\frac{1}{\bar{\xi}} - \frac{1}{\widehat{\xi}} \big) \log \frac{\widehat{c}}{\widehat{\xi}} + \frac{1}{\bar{\xi}} \log \frac{\widehat{\xi}}{\bar{\xi} }  \bigg) \\
=&  \bar{\upsilon} \bar{c} \mathbf{m}  \bigg( \big(\frac{1}{\bar{\xi}} - \frac{1}{\widehat{\xi}} \big) \log \frac{\widehat{c}}{\widehat{\xi}} + \frac{1}{\bar{\xi}} \log \frac{\widehat{\xi}}{\bar{\xi} }  \bigg).
\end{align*}
Here $\bar{\upsilon}= \sqrt{\frac{1}{\mathbf{m}}\sum_{s=1}^{\mathbf{m}}\upsilon_s^2}$, and $\bar{c}= \sqrt{\frac{1}{\mathbf{m}}\sum_{s=1}^{\mathbf{m}}(c^s)^2}$. Hence, the complexity of feature inclusion is given by 
\begin{align*}
T_{add} &= T_{a1}  + \varsigma np \mathbf{m} \\
&=  \bar{\upsilon} \bar{c} \mathbf{m}  \bigg( \big(\frac{1}{\bar{\xi}} - \frac{1}{\widehat{\xi}} \big) \log \frac{\widehat{c}}{\widehat{\xi}} + \frac{1}{\bar{\xi}} \log \frac{\widehat{\xi}}{\bar{\xi} }  \bigg) + \varsigma np \mathbf{m}.
\end{align*}

If all the active features are added to the active set $\mathcal{A}$, and the duality gap is small enough to stop feature inclusion, the value of $DoAdd$ becomes False in Algorithm~\ref{alg:imprv_primal_dual}. Next, we do the complexity analysis for both feature screening and accuracy pursuit stages. 

With $DoAdd=False$,  we do not need to use all the features from the original problem to compute the primal and dual objectives. In feature screening and accuracy pursuit stages, the duality gap in the outer loop can be computed using the features in active set $\mathcal{A}_s$, i.e., $\xi^s = \overline{DGap}$. Hence, the complexity of outer loop operations can be omitted for both feature screening and accuracy pursuit stages. 

 With $\xi^S$ as the duality gap  threshold of  feature screening, let $\breve{\xi} = \max \{\bar{\xi}, \xi^S\}$.  The complexity upper bound of feature screening is approximately given by 
\begin{align*}
T_{screen} &\leq  \upsilon_{\mathbf{m}} \bigg( \frac{c^{\mathbf{m}}}{\xi^S} \log \frac{c^{\mathbf{m}}}{\xi^S} - \frac{c^{\mathbf{m}}}{\breve{\xi}} \log \frac{c^{\mathbf{m}}}{\breve{\xi}} \bigg) \\
&=  \upsilon_{\mathbf{m}} \bigg( \big( \frac{c^{\mathbf{m}}}{\xi^S} - \frac{c^{\mathbf{m}}}{\breve{\xi}}\big) \log \frac{c^{\mathbf{m}}}{\breve{\xi}} + \frac{c^{\mathbf{m}}}{\xi^S} \log \frac{\xi^S}{\breve{\xi}}  \bigg) \\
&=  \upsilon_{\mathbf{m}} c^{\mathbf{m}} \bigg( \big( \frac{1}{\xi^S} - \frac{1}{\breve{\xi}}\big) \log \frac{c^{\mathbf{m}}}{\breve{\xi}} + \frac{1}{\xi^S} \log \frac{\xi^S}{\breve{\xi}}  \bigg) .
\end{align*}
If $\xi^S > \bar{\xi}$, it means feature inclusion and  feature screening can finish in the same  outer step, and we have $T_{screen} = 0$. 

Let $\widetilde{\xi} =\max \big\{\xi, \min \{\bar{\xi}, \xi^S\} \big\} $.  The complexity of the accuracy pursuit stage is given by 
\begin{align*}
T_{*} = & \upsilon_{*} \bigg( \frac{c^{*}}{\xi} \log \frac{c^{*}}{\xi} - \frac{c^{*}}{\widetilde{\xi}} \log \frac{c^{*}}{\widetilde{\xi}} \bigg) \\
= & \upsilon_{*} c^{*} \bigg( \big( \frac{1}{\xi} - \frac{1}{\widetilde{\xi}}\big) \log \frac{c^*}{\widetilde{\xi}} + \frac{1}{\xi} \log \frac{\widetilde{\xi}}{\xi}   \bigg) .
\end{align*}
If $\xi > \min \{\bar{\xi}, \xi^S\}$, we have $T_{*} =0$. 
Therefore, the complexity of Algorithm \ref{alg:imprv_primal_dual} can be written as: 
\begin{align*}
T =& T_{add} + T_{screen} + T_{*} \\
\leq &\bar{\upsilon} \bar{c} \mathbf{m}  \bigg( \big(\frac{1}{\bar{\xi}} - \frac{1}{\widehat{\xi}} \big) \log \frac{\widehat{c}}{\widehat{\xi}} + \frac{1}{\bar{\xi}} \log \frac{\widehat{\xi}}{\bar{\xi} }  \bigg) + \varsigma np \mathbf{m} \\
&+   \upsilon_{\mathbf{m}} c^{\mathbf{m}} \bigg( \big( \frac{1}{\xi^S} - \frac{1}{\breve{\xi}}\big) \log \frac{c^{\mathbf{m}}}{\breve{\xi}} + \frac{1}{\xi^S} \log \frac{\xi^S}{\breve{\xi}}  \bigg)   +  \upsilon_{*} c^{*} \bigg( \big( \frac{1}{\xi} - \frac{1}{\widetilde{\xi}}\big) \log \frac{c^*}{\widetilde{\xi}} + \frac{1}{\xi} \log \frac{\widetilde{\xi}}{\xi}   \bigg)  .
\end{align*}

Without loss of generality, let's assume that  only one variable is added to  $\mathcal{A}^*$ in each outer step, indicating $\mathbf{m} \leq |\mathcal{A}^*|$. Thus, 
\begin{align*}
T  \leq &\bar{\upsilon} \bar{c} |\mathcal{A}^*| \bigg( \big(\frac{1}{\bar{\xi}} - \frac{1}{\widehat{\xi}} \big) \log \frac{\widehat{c}}{\widehat{\xi}} + \frac{1}{\bar{\xi}} \log \frac{\widehat{\xi}}{\bar{\xi} }  \bigg) + \varsigma np |\mathcal{A}^*|\\
&+   \upsilon_{\mathbf{m}} c^{\mathbf{m}} \bigg( \big( \frac{1}{\xi^S} - \frac{1}{\breve{\xi}}\big) \log \frac{c^{\mathbf{m}}}{\breve{\xi}} + \frac{1}{\xi^S} \log \frac{\xi^S}{\breve{\xi}}  \bigg)   +  \upsilon_{*} c^{*} \bigg( \big( \frac{1}{\xi} - \frac{1}{\widetilde{\xi}}\big) \log \frac{c^*}{\widetilde{\xi}} + \frac{1}{\xi} \log \frac{\widetilde{\xi}}{\xi}   \bigg)  \\
=& \bar{\upsilon} \bar{c} |\mathcal{A}^*| \Omega\big(\bar{\xi}, \widehat{\xi}, \widehat{c}\big) + \varsigma np |\mathcal{A}^*| +   \upsilon_{\mathbf{m}} c^{\mathbf{m}}  \Omega\big(\xi^S, \breve{\xi}, c^{\mathbf{m}}\big) +  \upsilon_{*} c^{*} \Omega\big(\xi, \widetilde{\xi},  c^*\big).
\end{align*}

 With active set $\mathcal{A}_s$, the complexity of one iteration in the inner Algorithm~\ref{alg:primal_dual_inner} is  around $4n|\mathcal{A}_s|$, i.e., $\upsilon_s = 4n|\mathcal{A}_s|$. Therefore,
 \begin{align*}
T  \leq   4n \bar{c}\overline{|\mathcal{A}|} |\mathcal{A}^*| \Omega\big(\bar{\xi}, \widehat{\xi}, \widehat{c}\big) + 3 np |\mathcal{A}^*| +    4n c^{\mathbf{m}} |\mathcal{A}_{\mathbf{m}}| \Omega\big(\xi^S, \breve{\xi}, c^{\mathbf{m}}\big) +  4n c^{*}  |\mathcal{A}^*|\Omega\big(\xi, \widetilde{\xi},  c^*\big).
\end{align*}
Here $\overline{|\mathcal{A}|} = \sqrt{\frac{1}{\mathbf{m}}\sum_{s=1}^{\mathbf{m}}|\mathcal{A}_s|^2} $. It concludes that the complexity of Algorithm~\ref{alg:imprv_primal_dual} is  $O\bigg(  4n \bar{c}\overline{|\mathcal{A}|} |\mathcal{A}^*| \Omega\big(\bar{\xi}, \widehat{\xi}, \widehat{c}\big) + 3 np |\mathcal{A}^*| +    4n c^{\mathbf{m}} |\mathcal{A}_{\mathbf{m}}| \Omega\big(\xi^S, \breve{\xi}, c^{\mathbf{m}}\big) +  4n c^{*}  |\mathcal{A}^*|\Omega\big(\xi, \widetilde{\xi},  c^*\big) \bigg)$.
\end{proof}

\begin{remark}\label{rmk:outer_loop}
The maximum size  of the active set $\mathcal{A}_s$ is usually several times larger than $\mathcal{A}^*$, i.e., $|\mathcal{A}_s| = O(|\mathcal{A}^*|), \forall 1 \leq s \leq \mathbf{m}$. Then the complexity is simplified as  $O\bigg( \big( 4n \bar{c} |\mathcal{A}^*| \Omega (\bar{\xi}, \widehat{\xi}, \widehat{c}) + 3 np +    4n c^{\mathbf{m}} \Omega (\xi^S, \breve{\xi}, c^{\mathbf{m}} ) +  4n c^{*} \Omega (\xi, \widetilde{\xi},  c^*) \big) |\mathcal{A}^*| \bigg)$. In addition, if $\bar{\xi} < \min(\xi^S, \xi)$, we have $T_{screen} = T_{*}=0$,  and the complexity of Algorithm~\ref{alg:imprv_primal_dual} becomes $O\bigg( \big( 4n \bar{c} |\mathcal{A}^*| \Omega (\bar{\xi}, \widehat{\xi}, \widehat{c}) + 3 np \big) |\mathcal{A}^*| \bigg)$. In different setups, the proposed dynamic incremental strategy significantly reduces the redundant operations introduced by the inactive features.  
\end{remark}

\newpage

\section{Proof of Theorem~\ref{Thm:ball}}~\label{sec:proof_thm}

\noindent\textbf{Theorem~\ref{Thm:ball}} \  {\it
Assume that the primal loss functions $\{l_i(\cdot)\}_{i=1}^n$ are $1/\mu$-strongly smooth.  The range of the dual variable is  bounded via the duality gap value, i.e.,  $\forall \alpha \in \mathcal{F}^n, \beta \in \mathbb{R}^p$, $\{B(\alpha; r) : || \alpha -  \bar{\alpha}||_2  \leq r, r = \sqrt{ \frac{2(P(\beta) - D(\alpha)) } {\gamma}} \} $. 
Here $\gamma$ is a positive constant and $\gamma \geq \mu$. 
}

\begin{proof}
 With the assumption $l_i$ being $1/\mu$-smooth, its  conjugate function $l^*_i$ is $\mu$-strongly convex. 
With the dual problem 
\begin{align*} 
D(\alpha)& = -  \sum_{i=1}^n  l_i^*(\alpha_i)   + \sum_{j=1}^p \Psi(\eta_j(\alpha) ; \lambda_0, \lambda_1,  \lambda_2)  .
\end{align*} 
Here
\begin{align*} 
&\Psi(\eta_j(\alpha); \lambda_0,  \lambda_1, \lambda_2) 
 =   \begin{cases}
- \lambda_2 (|\eta_j(\alpha)| - \frac{\lambda_1}{2 \lambda_2})^2 + \lambda_0   &\text{if} \  \     |\eta_j(\alpha)|  \geq \eta_0 \\
0  &\text{if}  \  \   |\eta_j(\alpha)|  < \eta_0
\end{cases} .
\end{align*}
With $\eta_0 = \frac{ 2\sqrt{\lambda_0 \lambda_2} +  \lambda_1}{2\lambda_2}$ and  $\eta_j(\alpha) = -\frac{1}{2\lambda_2} x^{\top}_{\cdot j}  \alpha$, we can see that $\Psi(\eta_j(\alpha); \lambda_0,  \lambda_1, \lambda_2)$ is concave for $j, 1 \leq j \leq p$ regarding $\alpha$. Hence $D(\alpha)$ is concave. For given hyper-parameters $\lambda_0, \lambda_1, \lambda_2$, let $\Psi(\alpha)=\sum_{j=1}^p\Psi(\eta_j(\alpha); \lambda_0,  \lambda_1, \lambda_2)$, and $\Psi_j(\alpha)=\Psi(\eta_j(\alpha); \lambda_0,  \lambda_1, \lambda_2)$.  For $1 \leq j \leq p$,
\begin{align*}
\Psi_j'(\alpha) = & \begin{cases}
\mathrm{sign}(\eta_j(\alpha)) (|\eta_j(\alpha)| - \frac{\lambda_1}{2\lambda_2}) x_{\cdot j}  &\text{if} \  \     |\eta_j(\alpha)|  \geq \eta_0 \\
0  &\text{if}  \  \   |\eta_j(\alpha)|  < \eta_0
\end{cases} \\
=& \begin{cases}
\beta_j x_{\cdot j}  &\text{if} \  \     |\eta_j(\alpha)|  \geq \eta_0 \\
0  &\text{if}  \  \   |\eta_j(\alpha)|  < \eta_0
\end{cases}  .
\end{align*} 
Hence,
\begin{align*}
\Psi_j''(\alpha) = & \begin{cases}
-\frac{1}{2\lambda_2} x_{\cdot j}  x^{\top}_{\cdot j}  &\text{if} \  \     |\eta_j(\alpha)|  \geq \eta_0 \\
0  &\text{if}  \  \   |\eta_j(\alpha)|  < \eta_0
\end{cases}  .
\end{align*} 
Therefore, 
\begin{align*}
\Psi''(\alpha) =  -\frac{1}{2\lambda_2} \sum_{j\in S}x_{\cdot j}x^{\top}_{\cdot j}.
\end{align*} 
Here $S$ is the support set regarding $\alpha$, i.e. $S=\mathrm{supp}\big(\beta(\alpha)\big)=\{j \big| |\eta_j(\alpha)| \geq \eta_0\}=\{j \big| |x_{\cdot j}^{\top} \alpha| \geq 2 \lambda_2 \eta_0\}$.  
The smallest eigenvalue of Hessian matrix  $\Psi''(\alpha)$ depends on $X_S$. Let $\sigma_{min}(X_S)$ be the smallest eigenvalue of $X_S$, then $\Psi(\alpha)$ is concave, and it is also $\nu = \frac{\sigma_{min}(X_S)}{2 \lambda_2}$-strongly concave at point $\alpha$. 

 
With $l^*_i$ being $\mu$-strongly convex, $D(\alpha)$ is $\gamma$-strongly concave with $\gamma = \mu + \inf_{S} \frac{\sigma_{min}(X_S)}{2 \lambda_2} \geq \mu$. Then we have
\begin{align*}
D(\alpha_1) \leq  D(\alpha_2) + \langle\nabla_{\alpha} D(\alpha_2),  \alpha_1 - \alpha_2 \rangle - \frac{\gamma}{2} || \alpha_1 - \alpha_2 ||^2 .
\end{align*}
Let $\alpha_2 = \bar{\alpha}$, and $\alpha_1 = \alpha \in \mathcal{F}^n$. As $\bar{\alpha}$ maximizes $D(\alpha)$, $\langle\nabla_{\alpha} D(\bar{\alpha}),  \alpha - \bar{\alpha} \rangle \leq 0$. It implies
 \begin{align*}
D(\alpha) \leq  D(\bar{\alpha}) - \frac{\gamma}{2} || \alpha - \bar{\alpha} ||^2 .
\end{align*}
Thus we have a ball range for the dual variable
\begin{align*}
 \forall \alpha \in \mathcal{F}^n, \beta \in \mathbb{R}^p,  || \alpha -  \bar{\alpha}||_2  \leq \sqrt{ \frac{2(P(\beta) - D(\alpha)) } {\gamma}} =: r.
 \end{align*}
 This completes the proof.
\end{proof}

\newpage

\section{Inner Algorithm Analysis}\label{sec:supp_analysis}
In this section, we  present the complexity analysis of the inner updating Algorithm~\ref{alg:primal_dual_inner}. 

\subsection{Convergence of Inner Primal-dual Updating}
 We will show that under certain conditions $\beta(\alpha)$ is locally smooth around $\bar{\beta} = \beta(\bar{\alpha})$. For a given set of parameters $\lambda = \{\lambda_0, \lambda_1, \lambda_2\}$, $\beta(\alpha)$ corresponds to a set of support features $\mathrm{supp}(\beta(\alpha))$. We use $\bar{\delta}$ to represent the set in the dual feasible space with $\mathrm{supp}(\beta(\alpha)) = \mathrm{supp}(\bar{\beta})$.

\begin{lemma}\label{Lem:support}
Let $X = [x_1, ..., x_n]^T \in \mathbb{R}^{n \times p}$ be the data matrix,  $\eta_{j}(\bar{\alpha})$ be the $j$th entry of $\eta(\bar{\alpha})$, $S=\mathrm{supp}(\bar{\beta})$, and $N= \{j| \eta_j(\bar{\alpha}) = \eta_{0} \}$. Assume that $\{l_i\}_{i =1,..., n}$ are differentiable, and  let
\begin{align*}
\bar{\delta} =: 2 \lambda_2 \min \bigg\{\min_{j:j \in S} \frac{|\eta_j(\bar{\alpha})  |  -  \eta_{0}}{||x_{\cdot j}||}, 
\min_{j:j \in S^c \setminus N} \frac{ \eta_{0} - |\eta_j(\bar{\alpha})|}{||x_{\cdot j}||}\bigg\},
\end{align*}
with $|| \alpha - \bar{\alpha}|| \leq \bar{\delta}$, we have $\mathrm{supp}(\beta(\alpha)) = \mathrm{supp}(\bar{\beta})$, and $|| \beta(\alpha) - \bar{\beta}|| \leq \frac{\sigma_{max}(X_S)}{2\lambda_2} ||\alpha - \bar{\alpha}||$.
\end{lemma}

\begin{proof}
For any $\alpha $, we have
\begin{align}\label{eq:eta_alp}
\eta(\alpha) = - \frac{1}{2\lambda_2} \sum_{i=1}^n \alpha_i x_i  = - \frac{1}{2\lambda_2}  X^{\top}\alpha .
\end{align}
For a feature $j \in S= \mathrm{supp}(\bar{\beta})$,  we have $|\eta_j(\bar\alpha)| - \eta_{0} > 0 $, which is
\begin{align*}
& |\eta_j(\bar{\alpha}) + \eta_j(\alpha) - \eta_j(\bar{\alpha}))|  >  \eta_{0} .
\end{align*}
We try to find the space for $\alpha$s with the same support as $\bar{\alpha}$. We use the lower bound of the above inequality,
\begin{align*}
& |\eta_j(\bar{\alpha}) + \eta_j(\alpha) - \eta_j(\bar{\alpha}))| \geq |\eta_j(\bar{\alpha})  | - |\eta_j(\alpha) - \eta_j(\bar{\alpha})|   >  \eta_{0},
\end{align*}
yielding 
\begin{align*}
& |\eta_j(\alpha) - \eta_j(\bar{\alpha})| < |\eta_j(\bar{\alpha})|-\eta_{0}.
\end{align*}
With~\eqref{eq:eta_alp},
\begin{align*}
&  \frac{\big|\big| \alpha - \bar{\alpha} \big|\big|}{2\lambda_2} \big|\big| x_{\cdot j} \big|\big|   < |\eta_j(\bar{\alpha})  |  -  \eta_{0}.
\end{align*}
Hence
\begin{align*}
&\big|\big| \alpha - \bar{\alpha} \big|\big| < \min_{j:j \in S} \frac{2 \lambda_2 (|\eta_j(\bar{\alpha})  |  -  \eta_{0})}{||x_{\cdot j}||}.
\end{align*}
Similarly, for features $j$, $j \notin S $ and  $j \notin N$, 
\begin{align*}
&|\eta_j(\bar\alpha)|  < \eta_{0}, \end{align*}
yielding 
\begin{align*}
& |\eta_j(\alpha) - \eta_j(\bar{\alpha})| < \eta_{0} - |\eta_j(\bar{\alpha})|.
\end{align*}
With all $j$s, $j \in S^c \setminus N$
\begin{align*}
&\big|\big| \alpha - \bar{\alpha} \big|\big| < \min_{j:j \in S^c \setminus N} \frac{2 \lambda_2 (  \eta_{0} - |\eta_j(\bar{\alpha})|)}{||x_{\cdot j}||}. 
\end{align*}
Therefore, if $\alpha \in \bar{\delta}$, with 
\begin{align*}
|| \alpha - \bar{\alpha}|| \leq  \bar{\delta} = 2 \lambda_2 \min \bigg\{\min_{j:j \in S} \frac{|\eta_j(\bar{\alpha})  |  -  \eta_{0}}{||x_{\cdot j}||}, 
\min_{j:j \in S^c \setminus N} \frac{ \eta_{0} - |\eta_j(\bar{\alpha})|}{||x_{\cdot j}||}\bigg\}, 
\end{align*}
we have $\mathrm{supp}(\beta(\alpha)) = \mathrm{supp}(\bar{\beta})$. With $|| \alpha - \bar{\alpha}|| \leq \bar{\delta}$,  the primal problem  becomes a convex $\ell_1$ regularization problem without any redundant features. The super-gradient in Remark~\ref{rmk:sup-grad} becomes $g_{\alpha} = X_{S}\beta_S(\alpha) - l^{*'}(\alpha)$. With   $||g_{\alpha}||\rightarrow 0$,  $\beta_S(\alpha) = (X_{S}^{\top}X_{S})^{-1} X_{S}^{\top} l^{*'}(\alpha)$. As $S$ is fixed, with $||\alpha - \bar{\alpha}||$ being small, we have  $\mathrm{sign}(\beta(\alpha)) = \mathrm{sign}(\bar{\beta})$. It means
\begin{align*}
|| \beta(\alpha) - \bar{\beta}|| &= ||  \mathfrak{B}(\eta(\alpha)) -  \mathfrak{B}(\eta(\bar{\alpha}))|| \\
&\leq || \eta(\alpha) -  \eta(\bar{\alpha})|| = \frac{1}{2\lambda_2} || X_S(\alpha - \bar{\alpha})|| \\
&\leq \frac{\sigma_{max}(X_S)}{2\lambda_2} ||\alpha - \bar{\alpha}|| .
\end{align*}
This completes the proof of the lemma.
\end{proof}

Note that the above lemma can be extended to any pair of $\alpha_1, \alpha_2 \in \mathcal{F}^n$, and if they are close enough, they have the same support set. Let $\Psi( \alpha; \lambda_0, \lambda_1,  \lambda_2) = \sum_{j=1}^p \Psi(\eta_j(\alpha) ; \lambda_0, \lambda_1,  \lambda_2)  $, and it is easy to verify that  $\Psi( \alpha; \lambda_0, \lambda_1,  \lambda_2)$ is concave. 


\begin{lemma}\label{Lem:dual_range}
Assume that the primal loss functions $\{l_i()\}_{i=1}^n$ are $1/\mu$-strongly smooth. Then the following inequality holds for any $\alpha_1, \alpha_2 \in \mathcal{F}^n$,  and $g(\alpha_2) \in \partial D(\alpha_2)$:

 \begin{align*}
D(\alpha_1) \leq D(\alpha_2) +  \langle g_{\alpha_2}, \alpha_1 -   \alpha_2  \rangle -  \frac{  \gamma }{2 } ||  \alpha_1 - \alpha_2 ||^2.
 \end{align*}
 Moreover, $\forall \alpha \in \mathcal{F}^n$, and $g_{\alpha} \in \partial D(\alpha)$, $||\alpha - \bar{\alpha}|| \leq \sqrt{ \frac{2   \langle g_{\alpha}, \bar{ \alpha}- \alpha \rangle  }{ \gamma}} .$
 Here, $\gamma$ is the same in Theorem~\ref{Thm:ball}.
\end{lemma}
\begin{proof}
With the assumption $l_i$ being $1/\mu$-smooth, its  conjugate function $l^*_i$ is $\mu$-strongly convex. 
With the dual problem 
\begin{align*} 
D(\alpha)& = -  \sum_{i=1}^n  l_i^*(\alpha_i)   + \sum_{j=1}^p \Psi(\eta_j(\alpha) ; \lambda_0, \lambda_1,  \lambda_2)  .
\end{align*} 
Here
\begin{align*} 
&\Psi(\eta_j(\alpha); \lambda_0,  \lambda_1, \lambda_2) 
 =   \begin{cases}
- \lambda_2 (|\eta_j| - \frac{\lambda_1}{2 \lambda_2})^2 + \lambda_0   &\text{if} \  \     |\eta_j(\alpha)|  \geq \eta_0 \\
0  &\text{if}  \  \   |\eta_j(\alpha)|  < \eta_0
\end{cases} .
\end{align*}
According to the proof of Theorem~\ref{Thm:ball}, $\Psi(\alpha)= \sum_{j=1}^p \Psi(\eta_j(\alpha) ; \lambda_0, \lambda_1,  \lambda_2)$ is $\nu$-strongly concave with  $\nu = \inf_{\alpha} \nu(\alpha) = \frac{\sigma_{min}(X_{S_{\alpha}})}{2 \lambda}$, and $S_{\alpha} = \mathrm{supp}\big(\beta(\alpha)\big)$. 
When $S_{\alpha} = \emptyset$, we have $\nu(\alpha)=0$. 
 Now let us consider two arbitrary dual variables $\alpha_1, \alpha_2 \in \mathcal{F}^n$,
\begin{align*} 
\Psi(\alpha_1) \leq  \Psi(\alpha_2) + \Psi'(\alpha_2)^{\top}(\alpha_1 - \alpha_2)   -  \frac{\nu}{2}||\alpha_1  -  \alpha_2||^2   .
\end{align*} 
Hence,
\begin{align}\label{eq:dual1}
D(\alpha_1) = &-  \sum_{i=1}^n  l_i^*(\alpha_{1(i)})   + \sum_{j=1}^p \Psi(\eta_j(\alpha_1) ; \lambda_0, \lambda_1,  \lambda_2)  \notag \\
\leq & \sum_{i=1}^n\big( -l_i^*(\alpha_{2(i)}) - l_i^{*'}(\alpha_{2(i)}) (\alpha_{1(i)} - \alpha_{2(i)}) - \frac{\mu}{2}(\alpha_{1(i)}  - \alpha_{2(i)})^2\big) \notag \\
&+ \Psi(\alpha_2) + \Psi'(\alpha_2)^{\top}(\alpha_1 - \alpha_2)   - \frac{\nu}{2}||\alpha_1  - \alpha_2||^2   \notag \\
 \leq & D(\alpha_2) +  \langle g_{\alpha_2}, \alpha_1 -   \alpha_2  \rangle -  \frac{  \gamma }{2 } ||  \alpha_1 - \alpha_2 ||^2.
\end{align} 
Here $\alpha_{1(i)}$ is the $i$th entry of $\alpha_1$. This proves the first desirable inequality in the lemma. With the above inequality and using the fact $D(\alpha) \leq D(\bar{\alpha})$ we get that 
 \begin{align*}
D(\bar{\alpha})  \leq& D(\alpha) +  \langle g_{\alpha}, \bar{\alpha}-   \alpha \rangle -    \frac{  \gamma }{2 } ||  \alpha- \bar{\alpha} ||^2 
\leq   D(\bar{\alpha}) +  \langle g_{\alpha}, \bar{\alpha}-   \alpha \rangle -   \frac{  \gamma}{2 } ||  \alpha- \bar{\alpha} ||^2,
 \end{align*}
which leads to the second desired bound,
\begin{align*}
||\alpha - \bar{\alpha}|| \leq \sqrt{ \frac{2   \langle g_{\alpha}, \bar{ \alpha}- \alpha \rangle  }{ \mu + \nu(\alpha)}} .
\end{align*}
It concludes the proof of the lemma. 
\end{proof}

Different from the primal updating~\citep{Hazimeh18} or dual updating~\citep{Liu17} algorithms,   Algorithm~\ref{alg:primal_dual_inner} has both primal and dual updating steps.
Let $m_1 =\max_{j:1\leq j \leq p} |y^{\top}x_{\cdot j}|$, and $m_2 =\max_{j:1\leq j \leq p} ||X^{\top}x_{\cdot j}||$, $m_3 = \max_{i, \alpha_i^t \in \mathcal{F}} |l_i^{*'}(\alpha_i^t)|$, and  $\varrho = \sqrt{n}||\alpha^t||_{\infty} - \lambda_1$. We have the following theorem regarding the convergence of Algorithm~\ref{alg:primal_dual_inner}.\\

\begin{theorem}\label{Thm:inner_convg}
Assume that $l_i$ is $1/\mu$-smooth,  $||x_{i}|| \leq \vartheta  \ \forall 1\leq i \leq n$, and $||x_{\cdot j}||= 1  \ \forall 1\leq j \leq p$.  
By choosing $w_t = \frac{1}{t \gamma}$, then the sequence generated by Algorithm~\ref{alg:primal_dual_inner} satisfies the following estimation error inequality:
 \begin{align*}
|| \alpha^t - \bar{\alpha}||^2 \leq c_1\bigg( \frac{1}{t} +  \frac{\ln t}{t}  \bigg).
\end{align*}
Here $c_1 = \frac{   c_0^2}{ \mu ^2 } $, $ c_0=  \frac{ \sqrt{np} \vartheta }{2\lambda_2(1+2\lambda_2)}(2\lambda_2m_1 +  \varrho +  \sqrt{n}m_2 \varrho - 2\lambda_1\lambda_2)+\sqrt{n}m_3$, $\gamma$ is same as in Theorem~\ref{Thm:ball}. 
\end{theorem}


\begin{proof}
 Let us consider $g^t$, $g_i^t = x_i^{\top} \beta^t - l_i^{*'}(\alpha_i^t)$. After computing the primal $\beta^t$  with the primal-dual relation~\eqref{eq:bigB}, Algorithm~\ref{alg:primal_dual_inner} also performs primal coordinate descent starting with $\beta^t$ using~\eqref{eq:thresh} to the improve super-gradient $g^t$. 
 
 Let $\breve{\beta}^t$ be the output of operation~\eqref{eq:bigB} at step $t$.  
  From the expression of $\beta^t$~\eqref{eq:bigB}, if $\breve{\beta}_j^t \neq 0$, $\breve{\beta}^t_j (\alpha^t)=\mathrm{sign}\big(\eta_j(\alpha^t)\big) \big(|\eta_j(\alpha^t)| -\frac{\lambda_1 }{2\lambda_2} \big) $. With $||x_{\cdot j}|| = 1$, $|\eta_j(\alpha^t)|= \frac{|x_{\cdot j}^{\top}\alpha^t|}{2\lambda_2}\leq \frac{\sqrt{n}\varphi}{2\lambda_2}$. Here $\varphi = ||\alpha^t||_{\infty}$. Then we have 
\begin{align}\label{eq:bound_beta_1}
 |\breve{\beta}_j^t| \leq |\eta_j(\alpha^t)| - \frac{\lambda_1}{2\lambda_2} \leq \frac{\sqrt{n}\varphi - \lambda_1}{2\lambda_2}= \frac{\varrho}{2\lambda_2} , 
\end{align}
with  $\varrho = \sqrt{n}\varphi - \lambda_1$.

 According to~\eqref{eq:thresh}, with $\beta$ as the input,  the non-zero output at entry $j$
\begin{align*}
 \grave{\beta}_j = T(\beta; \lambda_0, \lambda_1,\lambda_2)= \mathrm{sign}(\tilde{\mathbf{\beta}}_j) \frac{|\tilde{\mathbf{\beta}}_j| - \lambda_1}{1 + 2\lambda_2},
\end{align*}
 with
\begin{align*}
\tilde{\mathbf{\beta}}_j = \big \langle y - \sum_{i:i\ne j, i\in S} x_{\cdot i}\mathbf{\beta}_i, x_{\cdot j}  \big\rangle 
= (y - X\mathbf{\beta})^{\top} x_{\cdot j} + \mathbf{\beta}_j  x_{\cdot  j}^{\top}x_{\cdot j} 
=y^{\top} x_{\cdot j} -  \beta^{\top}X^{\top} x_{\cdot j}  + \mathbf{\beta}_j 
\end{align*}
and $ S=\mathrm{supp}(\beta)$. With $\beta_j \neq 0$
we have $|\tilde{\mathbf{\beta}}_j| - \lambda_1 \geq 0$.
Then
 \begin{align*}
  |\grave{\beta}_j| &= \frac{1}{1+2\lambda_2} \big( | \tilde{\mathbf{\beta}}_j| -\lambda_1\big) \leq \frac{1}{1+2\lambda_2} \big( |y^{\top} x_{\cdot j}| +|  \beta^{\top}X^{\top} x_{\cdot j}|  + |\mathbf{\beta}_j|  -\lambda_1\big) \\
  &\leq \frac{1}{1+2\lambda_2} \big(  |y^{\top} x_{\cdot j}| +\sqrt{\beta^{\top}\beta  x_{\cdot j}^{\top}XX^{\top}x_{\cdot j} }+ |\beta_j| -\lambda_1\big) .
\end{align*}
 Let input $\beta = \breve{\beta}^t$, with~\eqref{eq:bound_beta_1},  the upper bound of the output after one round coordinate descent will be  
  \begin{align*}
  |\beta^t_j| & \leq \frac{1}{1+2\lambda_2} \big(  |y^{\top} x_{\cdot j}| +\sqrt{\breve{\beta}^{\top}\breve{\beta}  x_{\cdot j}^{\top}XX^{\top}x_{\cdot j} }+ |\breve{\beta}_j| -\lambda_1\big) \\
  &\leq   \frac{1}{1+2\lambda_2} \big(  |y^{\top} x_{\cdot j}| +  \frac{1 + \sqrt{n}||X^{\top}x_{\cdot j}||}{2\lambda_2}\varrho -\lambda_1\big) \\
  &\leq \frac{1}{2\lambda_2(1+2\lambda_2)}(2\lambda_2m_1 +  \varrho +  \sqrt{n}m_2 \varrho - 2\lambda_1\lambda_2):=\psi .
\end{align*}
 
Here $m_1 =\max_{j:1\leq j \leq p} |y^{\top}x_{\cdot j}|$, and $m_2 =\max_{j:1\leq j \leq p} ||X^{\top}x_{\cdot j}||$. 
 Let $m_3 = max_{i, \alpha_i^t \in \mathcal{F}} |l_i^{*'}(\alpha_i^t)|$. Then 
 \begin{align*}
  |g_i^t| \leq |x_i^{\top} \beta^t| + |l_i^{*'}(\alpha_i^t)|\leq  \sqrt{||x_i||^2||\beta^t||^2}+ m_3\leq   \sqrt{p}\psi \vartheta + m_3 \ , \quad  \forall 1\leq i \leq n.
\end{align*}
Hence,
 \begin{align}\label{eq:c0}
     ||g^t|| \leq c_0 := \sqrt{np}\psi \vartheta + \sqrt{n}m_3 .
 \end{align}

Let $h^t = ||\alpha^t - \bar{\alpha}||$ and $v^t = \langle g^t, \bar{\alpha} - \alpha^t \rangle$. The concavity of $D$ implies $v^t \geq 0$. According to Lemma~\ref{Lem:dual_range}, 
\begin{align}\label{eq:h_inequ}
h^t = ||\alpha^t - \bar{\alpha}|| \leq \sqrt{ \frac{2   v^t  }{ \gamma}} .
\end{align}

Let $\omega^t$ be the step size for dual variables at step $t$. Then we have 
\begin{align*}
(h^t)^2  =&||\alpha^t - \bar{\alpha}||^2\\
=& ||P_{\mathcal{F}^n}\big(\alpha^{t-1} + \omega^{t-1} g^{t-1} \big) - \bar{\alpha}||^2 \\
\leq & || \alpha^{t-1} +  \omega^{t-1} g^{t-1} - \bar{\alpha} ||^2  \\
=&  (h^{t-1})^2 - 2\omega^{t-1}v^{t-1} + (\omega^{t-1})^2 ||g^{t-1}||^2   \\
\leq & (h^{t-1})^2 - \omega^{t-1}(    \gamma ) (h^{t-1})^2 + (\omega^{t-1})^2c_0^2 . 
\end{align*}
The last step is due to~\eqref{eq:h_inequ} and~\eqref{eq:c0}. 
Let $\omega^{t-1} = \frac{ 1 }{   \gamma t} $. We get 
\begin{align*}
(h^t)^2  \leq \big(1 - \frac{1}{t} \big) (h^{t-1})^2 + \frac{c_0^2}{\big(  \gamma\big)^2 t^2} .
\end{align*}
Recursively applying the above inequality we get
\begin{align*}
(h^t)^2  \leq   \frac{c_0^2}{\gamma^2}  \bigg( \frac{1}{t} + \frac{\ln t}{t}\bigg) \leq  c_1 \bigg( \frac{1}{t} + \frac{\ln t}{t}\bigg).
\end{align*}
Here $c_0=  \frac{\vartheta \sqrt{np}}{2\lambda_2(1+2\lambda_2)}(2\lambda_2m_1 +  \varrho +  \sqrt{n}m_2 \varrho - 2\lambda_1\lambda_2)+\sqrt{n}m_3$.
This proves the bound in the theorem. 
\end{proof}

\newpage

\section{Duality}~\label{sec:proof_duality}
In this section, we provide the proofs to extend the duality theory~\citep{pilanci2015sparse,Liu17,yuan2020dual} to the generalized sparse learning problem~\eqref{eq:primal1}.  The derivation of duality presented here is  significantly different from  the duality of hard thresholding~\citep{Liu17,yuan2020dual} because the generalized problem  ~\eqref{eq:primal1} uses soft-regularization terms rather than hard constraints, and it also includes the combination of three regularization norms, i.e., $\ell_0$-, $\ell_1$-, and $\ell_2$-norms. We establish the  duality theory with the guarantee that the original non-convex problem  in~\eqref{eq:primal1} can be solved in the dual space.  

\subsection{Dual Problem}

Different from the sparse saddle point in~\cite{yuan2020dual}and~\cite{Liu17} requiring to be $k$-sparse regarding the primal variables, the saddle point in this paper can be taken as a general saddle point.

\begin{lemma} \label{Lem:lemma1}
For a given $\alpha \in \mathcal{F}^n$, let $\beta(\alpha) = \argmin_{\beta} L(\beta, \alpha) $. We have 
\begin{align*} 
\min_{\beta} L(\beta, \alpha)= -  \sum_{i=1}^n  l_i^*(\alpha_i)   + \sum_{j=1}^p \Psi(\eta_j(\alpha); \lambda_0, \lambda_1,  \lambda_2),
\end{align*} 
where 
\begin{align*} 
& \Psi(\eta_j(\alpha); \lambda_0,  \lambda_1, \lambda_2)   
=    \begin{cases}
- \lambda_2 (|\eta_j(\alpha)| - \frac{\lambda_1}{2 \lambda_2})^2 + \lambda_0   &\text{if} \  \     |\eta_j(\alpha)|  > \eta_0  \\
\big\{0, - \lambda_2 (|\eta_j(\alpha)| - \frac{\lambda_1}{2 \lambda_2})^2 + \lambda_0  \big\} \quad  &\text{if}  \  \    |\eta_j(\alpha)|  = \eta_0 \\
0  &\text{if}  \  \   |\eta_j(\alpha)|  < \eta_0
\end{cases} .
 \end{align*}
To be specific, $l^*$ is the conjugate function of $l$, $\eta(\alpha) := - \frac{1}{2\lambda_2} \sum_{i=1}^n \alpha_i x_i$ and 
 $\eta_0 := \frac{ 2\sqrt{\lambda_0 \lambda_2} +  \lambda_1}{2\lambda_2}$. The link function between $\alpha$ and $\beta$ is
\begin{align*} 
& ~~~~ \beta_j(\alpha)  =    \mathfrak{B}(\eta_j(\alpha))  = \begin{cases}
\mathrm{sign}\big(\eta_j(\alpha)\big) \big(|\eta_j(\alpha)| -\frac{\lambda_1 }{2\lambda_2} \big) &\text{if} \  \    |\eta_j(\alpha)|  > \eta_0 \\
\big\{0, \mathrm{sign}\big(\eta_j(\alpha)\big) \big(|\eta_j(\alpha)| -\frac{\lambda_1 }{2\lambda_2} \big) \big\}  &\text{if}  \  \  |\eta_j(\alpha)|  = \eta_0  \\
0  &\text{if}  \  \   |\eta_j(\alpha)|  < \eta_0
\end{cases}
.
\end{align*}
\end{lemma}

\begin{proof}
 With $l^*$ as the conjugate of $l$, the primal problem can be rewritten as 
\begin{align*} 
&\min_{\beta \in \mathbb{R}^p}  L(\beta, \alpha) \\
=& \min_{\beta \in \mathbb{R}^p}   \sum_{i=1}^n   \big(\alpha_i \beta^{\top}x_i - l_i^*(\alpha_i) \big) + \lambda_0 ||\beta||_0 + \lambda_1 ||\beta||_1 + \lambda_2 ||\beta||^2_2\\
=&-  \sum_{i=1}^n  l_i^*(\alpha_i)  + \min_{\beta \in \mathbb{R}^p}  \sum_{i=1}^n   \big(\alpha_i \beta^{\top}x_i \big) + \lambda_0 ||\beta||_0 + \lambda_1 ||\beta||_1 + \lambda_2 ||\beta||^2_2 \\
=&  -  \sum_{i=1}^n  l_i^*(\alpha_i)   +  \min_{\beta \in \mathbb{R}^p} \sum_{j=1}^p \big(\sum_{i=1}^n \alpha_i x_{ij} \big)\beta_j  + \lambda_0 ||\beta||_0 + \lambda_1 ||\beta||_1 + \lambda_2 ||\beta||^2_2 \\
=& -  \sum_{i=1}^n  l_i^*(\alpha_i)   + \sum_{j=1}^p \Phi( \sum_{i=1}^n \alpha_i x_{ij}; \lambda_0, \lambda_1,  \lambda_2).
\end{align*}

Let $\tau_j = \alpha^{\top}  x_{\cdot j}$. We have 
\begin{align*} 
 \Phi(\tau_j; \lambda_0, \lambda_1, \lambda_2) =&  \min_{\beta_j}    \tau_j \beta_j + \lambda_0 \mathbf{I} \{\beta_j \neq 0 \}+ \lambda_1|\beta_j|+ \lambda_2 \beta_j^2 \\
 =& \min_{\beta_j}   \lambda_2 \bigg( \beta_j + \frac{\tau_j}{2\lambda_2}\bigg)^2  +   \lambda_0 \mathbf{I} \{\beta_j \neq 0 \}+ \lambda_1|\beta_j| - \frac{\tau_j^2}{4 \lambda_2} .
\end{align*}
Let
\begin{align*} 
h(u) =   \lambda_2 \bigg( u + \frac{\tau_j}{2\lambda_2}\bigg)^2  +   \lambda_0 \mathbf{I} \{u \neq 0 \}+ \lambda_1|u| - \frac{\tau_j^2}{4 \lambda_2} .
\end{align*}

If $u \neq 0$, we have the soft-thresholding solution:  
\begin{align*} 
\hat{u} = -\frac{1}{2\lambda_2} \mathrm{sign}(\tau_j)\max(|\tau_j| - \lambda_1,0) .
\end{align*}

With $|\tau_j| > \lambda_1$,  
\begin{align*} 
h(\hat{u}) &=  \lambda_2 \bigg( -\frac{\tau_j - \lambda_1 \mathrm{sign}(\tau_j) }{2\lambda_2}  + \frac{\tau_j}{2\lambda_2}\bigg)^2  +   \lambda_0 +\frac{ \lambda_1}{2\lambda_2}(|\tau_j| - \lambda_1) - \frac{\tau_j^2}{4 \lambda^2_2}  \\
&= \lambda_0 + \frac{\lambda_1 |\tau_j|}{2\lambda_2} - \frac{\lambda_1^2}{4\lambda_2} -  \frac{\tau_j^2}{4 \lambda_2} \\
&=  \lambda_0  - \frac{1}{4\lambda_2} \big( \lambda_1^2 + \tau_j^2 - 2\lambda_1|\tau_j|\big) \\
&=  \lambda_0  -   \frac{1}{4\lambda_2} \big( |\tau_j| -  \lambda_1 \big) ^2 .
\end{align*}
With $h(\hat{u}) < h(0)=0$, we get 
\begin{align*} 
 \lambda_0  <   \frac{1}{4\lambda_2} \big( |\tau_j| -  \lambda_1 \big) ^2  \implies   |\tau_j|  > 2\sqrt{\lambda_0 \lambda_2} +  \lambda_1 .
\end{align*}
Hence, $\hat{u}$ is the minimizer when $ |\tau_j|  > 2\sqrt{\lambda_0 \lambda_2} +  \lambda_1$.  We have
\begin{align*} 
 \Phi(\tau_j; \lambda_0, \lambda_1, \lambda_2) =  -   \frac{1}{4\lambda_2} \big( |\tau_j| -  \lambda_1 \big) ^2 +  \lambda_0  .
\end{align*}

When $2\sqrt{\lambda_0 \lambda_2} +  \lambda_1  = |\tau_j|$, both $\hat{u}$ and 0 are 
minimizers. Then 
\begin{align*} 
 \Phi(\tau_j; \lambda_0, \lambda_1, \lambda_2)  = \{ 0, -\frac{1}{4\lambda_2} (|\tau_j| - \lambda_1)^2 + \lambda_0 \}.
\end{align*}

 If $2\sqrt{\lambda_0 \lambda_2} +  \lambda_1  > |\tau_j| $, 0 is the minimizer, and $ \Psi(\tau_j, \lambda_0, \lambda_1, \lambda_2)  = 0$.

 The optimal primal can be written as 

\begin{align*} 
\beta^*_j & =    \mathfrak{A}(\tau_j; \lambda_0, \lambda_1, \lambda_2)  =  \begin{cases}
\mathrm{sign}(\tau_j) \frac{\lambda_1 - |\tau_j|}{2\lambda_2} &\text{if} \  \    |\tau_j|  > 2\sqrt{\lambda_0 \lambda_2} +  \lambda_1 \\
\{0,  \mathrm{sign}(\tau_j) \frac{\lambda_1 - |\tau_j|}{2\lambda_2}  \} \quad  &\text{if}  \  \   |\tau_j|  = 2\sqrt{\lambda_0 \lambda_2} +  \lambda_1 \\
0  &\text{if}  \  \   |\tau_j|   < 2\sqrt{\lambda_0 \lambda_2} +  \lambda_1
\end{cases} \\
&= \mathfrak{B}(\eta_j(\alpha)) .
\end{align*}
Here $\eta_0 = \frac{ 2\sqrt{\lambda_0 \lambda_2} +  \lambda_1}{2\lambda_2} $.
With the optimal $\beta(\alpha)$, $L(\beta, \alpha)$ can be written as 
\begin{align*} 
 \Phi(\tau_j; \lambda_0,  \lambda_1, \lambda_2)& =   \begin{cases}
-\frac{1}{4\lambda_2} (|\tau_j| - \lambda_1)^2  + \lambda_0   &\text{if} \  \    |\tau_j|  > 2\sqrt{\lambda_0 \lambda_2} +  \lambda_1 \\
\{0,  -\frac{1}{4\lambda_2} (|\tau_j| - \lambda_1)^2 + \lambda_0   \} \quad  &\text{if}  \  \   |\tau_j|  = 2\sqrt{\lambda_0 \lambda_2} +  \lambda_1 \\
0  &\text{if}  \  \   |\tau_j|   < 2\sqrt{\lambda_0 \lambda_2} +  \lambda_1
\end{cases} .
\end{align*}
Alternatively, 
\begin{align*} 
 \Psi(\eta_j(\alpha); \lambda_0,  \lambda_1, \lambda_2)   
&= \Phi(\tau_j; \lambda_0,  \lambda_1, \lambda_2)  \\
 & =   \begin{cases}
- \lambda_2 (|\eta_j(\alpha)| - \frac{\lambda_1}{2 \lambda_2})^2 + \lambda_0   &\text{if} \  \     |\eta_j(\alpha)|  >  \eta_0 \\
\big\{0, - \lambda_2 (|\eta_j(\alpha)| - \frac{\lambda_1}{2 \lambda_2})^2 + \lambda_0  \big\} \quad  &\text{if}  \  \    |\eta_j(\alpha)|  = \eta_0 \\
0  &\text{if}  \  \   |\eta_j(\alpha)|  < \eta_0
\end{cases} .
\end{align*}
Here $\eta(\alpha) = - \frac{1}{2\lambda_2} \sum_{i=1}^n \alpha_i x_i$, and $\eta_0 = \frac{ 2\sqrt{\lambda_0 \lambda_2} +  \lambda_1}{2\lambda_2} $. It concludes the proof. 
\end{proof}

\begin{lemma} \label{Lem:saddle_point}
 (Saddle Point). Let $\bar{\beta}$ be a primal vector and $\bar{\alpha}$ a  dual vector. Then $ (\bar{\alpha}, \bar{\beta})$ is a   saddle point of $L$ if and only if the following conditions hold:
 
 a) $\bar{\beta}$ solves the primal problem;
 
 b) $\bar{\alpha} \in [\partial l_1(\bar{\beta}^{\top}x_1),  \partial l_2(\bar{\beta}^{\top}x_2), ..., \partial l_n(\bar{\beta}^{\top}x_n)]^{\top}$;
 
 c) $\bar{\beta}_j = \mathfrak{B}(\eta_j(\bar{\alpha}))$.
\end{lemma}

\begin{proof}
$\Leftarrow$:  If the pair $(\bar{\beta}, \bar{\alpha})$ is a  saddle point for $L$, then from the definition of conjugate convexity and inequality in the definition of  saddle point we have 
\begin{align*} 
P(\bar{\beta}) = \max_{\alpha \in \mathcal{F}^n} L(\bar{\beta}, \alpha) \leq  L(\bar{\beta}, \bar{\alpha})  \leq \min_{\beta \in \mathbb{R}^p } L(\beta, \bar{\alpha}) .
 \end{align*}
On the other hand, we know that for any $\beta \in \mathbb{R}^p$ and $\alpha \in \mathcal{F}^n$
\begin{align*} 
L(\beta, \alpha) \leq \max_{\alpha' \in \mathcal{F}^n} L(\beta, \alpha') = P(\beta).
 \end{align*}
By combining the preceding two inequalities we obtain
\begin{align*} 
P(\bar{\beta}) \leq \min_{\beta \in \mathbb{R}^p} L(\beta, \bar{\alpha}) \leq \min_{\beta \in \mathbb{R}^p} P(\beta) \leq P(\bar{\beta}).
 \end{align*}
Therefore $P(\bar{\beta}) = \min_{ \beta \in \mathbb{R}^p } P(\beta)$, i.e., $\bar{\beta}$ solves the primal problem, which proves the necessary condition~a). Moreover, the above arguments lead to 
\begin{align*} 
P(\bar{\beta}) = \max_{\alpha  \in \mathcal{F}^n } L(\bar{\beta}, \alpha) = L(\bar{\beta}, \bar{\alpha}) .
 \end{align*}
Then from the maximizing argument property of the convex conjugate we have $\bar{\alpha}_i \in \partial l_i(\bar{\beta}^{\top} x_i) $, and it leads to condition b).
Note that 
\begin{align} \label{eq:L_Thm1}
L(\beta, \bar{\alpha}) = \lambda_2\bigg|\bigg| \beta + \frac{1}{ 2\lambda_2} \sum_{i=1}^n \bar{\alpha}_i x_i \bigg|\bigg|^2 -   \sum_{i=1}^n l_i^*(\bar{\alpha}_i) + \lambda_1||\beta||_1 + \lambda_0 ||\beta||_0+C .
\end{align}
Let $\bar{F} = \mathrm{supp}(\bar{\beta})$. Since the above analysis implies $L(\bar{\beta}, \bar{\alpha})=\min_{\beta} L(\beta, \bar{\alpha})$, with $\eta(\bar{\alpha}) = - \frac{1}{2\lambda_2} \sum_{i=1}^n \bar{\alpha}_i x_i$, it must hold that (more details refer to the proof of Lemma~\ref{Lem:lemma1})
$\bar{\beta}_j = \mathfrak{B}(\eta_j(\bar{\alpha}))$.
This validates the condition c).

$\Rightarrow$: Inversely, let us assume that $\bar{\beta}$ is a solution to the primal problem~(condition a)), and $\bar{\alpha}_i \in \partial l_i(\bar{\beta}^{\top}x_i)$~(condition b)). Again from the maximizing argument property of the convex conjugate we know that $l_i(\bar{\beta}^{\top} x_i) = \bar{\alpha} \bar{\beta}^{\top}x_i - l_i^*(\bar{\alpha}_i)$. This leads to 
\begin{align} 
L(\bar{\beta}, \alpha) \leq  P(\bar{\beta}) = \max_{\alpha \in \mathcal{F}^n} L(\bar{\beta}, \alpha) = \ L(\bar{\beta}, \bar{\alpha}). \label{eq:ThOne1_2}
 \end{align}
The sufficient condition~(c) guarantees that based on the expression of \eqref{eq:L_Thm1}, for any $\beta$, we have 
\begin{align} 
L(\bar{\beta}, \bar{\alpha}) \leq  L(\beta, \bar{\alpha}) \label{eq:ThOne2}.
\end{align}
By combining the inequalities~\eqref{eq:ThOne1_2} and~\eqref{eq:ThOne2} we get that for any $\beta$ and $\alpha$ 
 \begin{align*}
L(\bar{\beta}, \alpha) \leq L(\bar{\beta}, \bar{\alpha})  \leq L( \beta, \bar{\alpha}) .
\end{align*}
This shows that $(\bar{\beta}, \bar{\alpha})$ is a  saddle point of the Lagrangian $L$. 
\end{proof}

\begin{theorem}\label{Thm:minimax}
Let $\bar{\beta} \in \mathbb{R}^p $ be a primal vector and $\bar{\alpha} \in \mathcal{F}^n$  regarding L, then
\vspace{-0.1in}

\begin{enumerate} 
\item $ (\bar{\alpha}, \bar{\beta})$ is a  saddle point of $L$ if and only if the following conditions hold:
\begin{enumerate}
\item $\bar{\beta}$ solves the primal problem; 
 
\item  $\bar{\alpha} \in [\partial l_1(\bar{\beta}^{\top}x_1),  \partial l_2(\bar{\beta}^{\top}x_2), ..., \partial l_n(\bar{\beta}^{\top}x_n)]^{\top}$;
\item
$\bar{\beta}_j =   \mathfrak{B}(\eta_j(\bar{\alpha}))$.
\end{enumerate}

 \item The mini-max relationship 
\begin{align} \label{eq:th2}
\max_{\alpha \in \mathcal{F}^n} \min_{\beta} L(\beta, \alpha) = \min_{\beta} \max_{\alpha \in  \mathcal{F}^n} L(\beta, \alpha) 
\end{align}
holds if and only if there exists a saddle point $(\bar{\beta}, \bar{\alpha})$ for $L$. 

\item The corresponding dual problem of~\eqref{eq:primal1} is  written as
\begin{align} \label{eq:dual_1}
  &\max_{\alpha \in \mathcal{F}^n} D(\alpha) =  \max_{\alpha \in \mathcal{F}^n}-  \sum_{i=1}^n  l_i^*(\alpha_i)   + \sum_{j=1}^p \Psi(\eta_j(\alpha); \lambda_0,  \lambda_1, \lambda_2),
\end{align} 
where  $l^*$ is the conjugate function of $l$.  The primal dual link is written as $\beta_j(\alpha)  =  \mathfrak{B}(\eta_j(\alpha))$.

\item (Strong duality) $\bar{\alpha}$ solves the dual problem in~\eqref{eq:dual}, i.e., $D(\bar{\alpha}) \geq D(\alpha), \alpha \in \mathcal{F}^n$, and $P(\bar{\beta}) = D(\bar{\alpha})$ if and only if the pair $(\bar{\beta}, \bar{\alpha})$ satisfies the  three conditions given by (a)$\sim$(c).
\end{enumerate}

\end{theorem}

\begin{proof}
According to Lemma~\ref{Lem:saddle_point}, statement-1 can be proved. We focus on statements 2-4. The following Part (I),  Part (II), and Part (III) are proofs of statement-2, statement-3, and statement-4, respectively. 

Part (I): the mini-max relationship in  statement-2.

$\Rightarrow$:
Let $(\bar{\beta}, \bar{\alpha})$ be a  saddle point for $L$. On one hand, note that the following holds for any $\beta'$ and $\alpha'$,
\begin{align*} 
\min_{\beta} L(\beta, \alpha') \leq  L(\beta', \alpha')  \leq \max_{\alpha \in \mathcal{F}^n} L(\beta', \alpha), 
\end{align*}
which implies
\begin{align} 
\max_{\alpha \in \mathcal{F}^n }\min_{\beta} L(\beta, \alpha)  \leq \min_{\beta} \max_{\alpha \in \mathcal{F}^n} L(\beta, \alpha) . \label{eq:th2_1}
\end{align}
On the other hand, since $(\bar{\beta}, \bar{\alpha})$ is a  saddle point for $L$, the following is true:
\begin{align}
& ~ \max_{\alpha \in \mathcal{F}^n}\min_{\beta} L(\beta, \alpha)  \leq  \max_{\alpha \in \mathcal{F}^n } L(\bar{\beta}, \alpha)  \nonumber \\
& \leq  \max_{\alpha \in \mathcal{F}^n} L(\bar{\beta},  \bar{\alpha})  \leq \min_{\beta} L(\beta,\bar{ \alpha}) 
 \leq \max_{\alpha \in \mathcal{F}^n }\min_{\beta} L(\beta, \alpha) \label{eq:th2_2} .
\end{align}
By combining~\eqref{eq:th2_1} and~\eqref{eq:th2_2}, we prove the equality~\eqref{eq:th2}. 

$\Leftarrow$: 
Assume that the equality in~\eqref{eq:th2} holds. Let us define $\bar{\beta}$ and $\bar{\alpha}$ such that 
\[\max_{\alpha \in \mathcal{F}^n} L(\bar{\beta}, \alpha) = \min_{\beta}  \max_{\alpha \in \mathcal{F}^n } L(\beta, \alpha) \] 
and 
\[\min_{\beta} L(\beta, \bar{\alpha}) = \max_{\alpha \in \mathcal{F}^n}  \min_{\beta}  L(\beta, \alpha) .
\]
Then we can see that for any $\alpha$, $L(\bar{\beta}, \bar{\alpha}) \geq \min_{\beta} L(\beta, \bar{\alpha}) = \max_{\alpha' \in \mathcal{F}^n} L(\bar{\beta}, \alpha')$, by~\eqref{eq:th2}. 
In the meantime, for any $\beta$
 \begin{align*} 
L(\bar{\beta}, \bar{\alpha}) \leq \max_{\alpha \in \mathcal{F}^n} L(\bar{\beta}, \alpha) =  \min_{\beta} L(\beta', \bar{\alpha}) \leq L(\beta, \bar{\alpha}).
\end{align*}
This shows that $(\bar{\beta}, \bar{\alpha})$ is a  saddle point for L.

Part (II): the dual form in statement-3.

According to Lemma~\ref{Lem:lemma1}, for any $\alpha$, the minimizing $\beta$ for $L(\beta, \alpha)$ satisfies: 
\begin{align} \label{eq:beta_j}
\beta_j (\alpha)  =    \mathfrak{B}(\eta_j(\alpha))  = \begin{cases}
\mathrm{sign}\big(\eta_j(\alpha)\big) \big(|\eta_j(\alpha)| -\frac{\lambda_1 }{2\lambda_2} \big) &\text{if} \  \    |\eta_j(\alpha)|  > \eta_0 \\
\big\{0, \mathrm{sign}\big(\eta_j(\alpha)\big) \big(|\eta_j(\alpha)| -\frac{\lambda_1 }{2\lambda_2} \big) \big\} \quad  &\text{if}  \  \  |\eta_j(\alpha)|  = \eta_0  \\
0  &\text{if}  \  \   |\eta_j(\alpha)|  < \eta_0
\end{cases}
.
\end{align}

Then we have 
\begin{align} \label{eq:dual1}
D(\alpha) = -  \sum_{i=1}^n  l_i^*(\alpha_i)   + \sum_{j=1}^p \Psi(\eta_j ; \lambda_0, \lambda_1,  \lambda_2),
\end{align} 
where 
\begin{align} \notag
 \Psi(\eta_j(\alpha); \lambda_0,  \lambda_1, \lambda_2) &=   \begin{cases}
- \lambda_2 (|\eta_j(\alpha)| - \frac{\lambda_1}{2 \lambda_2})^2 + \lambda_0   &\text{if} \  \     |\eta_j(\alpha)|  > \eta_0  \\
\big\{0, - \lambda_2 (|\eta_j(\alpha)| - \frac{\lambda_1}{2 \lambda_2})^2 + \lambda_0  \big\} \quad  &\text{if}  \  \    |\eta_j(\alpha)|  = \eta_0 \\
0  &\text{if}  \  \   |\eta_j(\alpha)|  < \eta_0
\end{cases} \\ \label{eq:dual_normal}
&= - \lambda_2 ||\beta(\alpha)||^2_2 + \lambda_0 ||\beta(\alpha)||_0 .
 \end{align}
Here $\eta_0 = \frac{ 2\sqrt{\lambda_0 \lambda_2} +  \lambda_1}{2\lambda_2} $. Assume that we have two arbitrary dual variables $\alpha_1, \alpha_2 \in \mathcal{F}^n$ and any $g(\alpha_2) \in [\beta(\alpha_2)^{\top} x_1 - {l_1^*}^{'}(\alpha_{2(1)}), ..., \beta(\alpha_2)^{\top} x_n - {l_n^*}^{'}(\alpha_{2(n)})] $. Here $\alpha_{2(n)}$ is the $n$th entry of $\alpha_{2}$. $L(\beta, \alpha)$ is concave in terms of $\alpha$ given any fixed $\beta$. According to the definition of $D(\alpha)$, we have 
 \begin{align*}
 D(\alpha_1) = L(\beta(\alpha_1), \alpha_1) \leq L(\beta(\alpha_2), \alpha_1)  \leq L(\beta(\alpha_2), \alpha_2)  +\langle g(\alpha_2) , \alpha_1 - \alpha_2 \rangle .
 \end{align*}
Hence $D(\alpha)$ is concave and the super gradient is as given. 

Part (III): Strong duality.
$\Rightarrow$:
Given the conditions a)-c), we can see that the pair~($\bar{\beta}$, $\bar{\alpha}$) forms a  saddle point of $L$. Thus based on the definitions of  saddle 
point and dual function  $D(\alpha)$, we can show that 
\begin{align*} 
D(\bar{\alpha}) = \min_{\beta} L(\beta, \bar{\alpha}) \geq  L(\bar{\beta}, \bar{\alpha}) \geq  L(\bar{\beta}, \alpha) \geq  D(\alpha) .
\end{align*}
This implies that $\bar{\alpha}$ solves the dual problem. Furthermore, Theorem~\ref{Thm:minimax}-2 guarantees the following 
\begin{align*} 
D(\bar{\alpha}) =\max_{\alpha \in \mathcal{F}^n} \min_{\beta} L(\beta, \alpha) =\min_{\beta} \max_{\alpha \in \mathcal{F}^n}  L(\beta, \alpha) = P(\bar{\beta}) . 
\end{align*}
This indicates that the primal and dual optimal values are equal to each other.

$\Leftarrow$: 
Assume that $\bar{\alpha}$ solves the dual problem and $D(\bar{\alpha}) =  P(\bar{\beta}) $. Since $D(\bar{\alpha}) \leq  P(\beta) $ holds for any $\beta$, $\bar{\beta}$ must be the sparse
minimizer of $P(\beta)$. It follows that 
\begin{align*} 
\max_{\alpha \in \mathcal{F}^n} \min_{\beta} L(\beta, \alpha) = D(\bar{\alpha})  = P(\bar{\beta}) =\min_{\beta} \max_{\alpha \in \mathcal{F}^n}  L(\beta, \alpha) . 
\end{align*}
From the $\Leftarrow$ argument in the proof of Theorem~\ref{Thm:minimax}-2, we get that conditions a)-c)in Theorem~\ref{Thm:minimax}-1 should be satisfied for $(\bar{\beta}, \bar{\alpha})$. This completes the proof.
\end{proof}

Compared to the dual problem developed in~\cite{Liu17,yuan2020dual} regarding  hard thresholding, the soft thresholding in \eqref{eq:beta_j} corresponds to the combination of $\ell_1$ and $\ell_0$ penalties, which typically lead to better results on datasets with low SNR values~\citep{pmlr-v65-david17a,mazumder2022subset}. \\

\subsection{Analysis on Strong Duality}

 Theorem~\ref{Thm:minimax}-1 gives the sufficient and necessary conditions for the existence of a  saddle point of the Lagrangian $L$~\eqref{eq:lagrangian}.  Theorem~\ref{Thm:minimax}-2 is on the min-max side of the problem, and it  provides conditions under which one can exchange min-max to max-min regarding $L$~\eqref{eq:lagrangian}. 
\begin{remark}
Applying Theorem~\ref{Thm:minimax}, the following mini-max relationship 
\begin{align} \label{eq:rmk1}
\max_{\alpha \in \mathcal{F}^n} \min_{\beta} L(\beta, \alpha) = \min_{\beta} \max_{\alpha \in  \mathcal{F}^n} L(\beta, \alpha) 
\end{align}
holds if and only if there exists a primal vector $\bar{\beta} \in \mathbb{R}^p$ and a dual vector $\bar{\alpha} \in \mathcal{F}^n$ such that conditions (a) $\sim$  (c) in Theorem\ref{Thm:minimax}-1 are satisfied.
Moreover, by calculations, it can be checked that \eqref{eq:rmk1} holds automatically for $ l(\cdot)$ being the square loss function.
\end{remark}

We use $P(\beta) = F(\beta)$ to represent the primal objective, and $D(\alpha)$ for the dual objective given in~(\ref{eq:dual}). 
Theorem~\ref{Thm:minimax}-3 indicates that the dual objective function $D(\alpha)$ is concave and the following remark explicitly gives the expression of its super-differential.
\begin{remark}\label{rmk:sup-grad}
The super-differential of  the dual form~\eqref{eq:dual} at $\alpha$ is given by 
$\nabla D(\alpha) = X\beta(\alpha) - l^{*'}(\alpha) = [\beta(\alpha)^{\top} x_1 -   {l^{*}_1}^{'}(\alpha_1), ..., \beta(\alpha)^{\top} x_n - {l^{*}_n}^{'}(\alpha_n )]^{\top}$.
\end{remark}

The super-gradient can be alternatively derived through the partial derivative of the Lagrangian $L$~\eqref{eq:lagrangian} regarding $\alpha$. The sparse strong duality theory in Theorem~\ref{Thm:minimax}-4 gives the sufficient and necessary conditions under which the optimal values of the primal and dual problems~coincide. 
According to Theorem~\ref{Thm:minimax}-4, the  primal-dual gap reaches zero at the primal-dual pair $(\beta, \alpha)$ if and only if the conditions (a) $\sim$ (c) in Theorem~\ref{Thm:minimax}-1 hold. The  duality theory developed in this section suggests a natural way for finding the global minimum of the sparsity-constrained minimization problem in~\eqref{eq:primal1} via  primal-dual optimization methods. Let $(l^{*'})^{-1}$ be the inverse of $l^{*'}$, we have the following remark with $0 \in \nabla D(\bar{\alpha})$ at $\bar{\alpha}$.
\begin{remark}\label{rmk:dual_condt}
If $(\bar{\beta}, \bar{\alpha})$ satisfies the conditions in Theorem~\ref{Thm:minimax}-1, we have $\bar{\alpha} \in (l^{*'})^{-1}(X\bar{\beta}) \cap \mathcal{F}^n$. 
\end{remark}

Strong duality holds when both the primal and dual variables reach the optimal values. Before attaining the optimal values, the duality gap value can be bounded  by the current dual variable estimations. The closer the current estimation $\alpha$ and the  optimal value $\bar{\alpha}$ are, the smaller the duality gap will be. As long as $\beta$ reaches its optimal value, all the conditions in Theorem~\ref{Thm:minimax}-1 can be met, and it is because the dual problem is concave.

Different from~\cite{yuan2020dual,Liu17}, a generalized sparse problem is studied in this paper. The methodology developed here can be easily extended to plain $\ell_1$ or $\ell_0$ problems (with  the $\ell_2$ term),  group sparse structures or fused sparse structures,  or even more  complex and  mixed sparse structures that we cannot or do not need to specify the active feature number $k$ value as in~\eqref{eq:hard_k}.


\end{document}